\RequirePackage[svgnames]{xcolor}

\documentclass[11pt,letterpaper]{mystyle}

\usepackage[all]{hypcap}
\usepackage[svgnames]{xcolor}
\usepackage[comma,authoryear,compress]{natbib}
\usepackage{hyperref}[citecolor=lightblue]

\hypersetup{
    colorlinks = true,
    citecolor = {YaleBlue},
}

\usepackage{algorithm}
\usepackage{algorithmicx}
\usepackage{algpseudocode}
\usepackage{microtype}
\usepackage{graphicx}
\expandafter\def\csname ver@subfig.sty\endcsname{}
\usepackage{booktabs} %
\usepackage{float}
\usepackage{bigstrut}

\usepackage{amsmath}
\usepackage{amssymb}
\usepackage{mathtools}
\usepackage{amsthm}
\usepackage{multirow}
\usepackage{mathrsfs}
\usepackage{nicefrac}
\usepackage{dsfont}
\usepackage{enumitem}
\usepackage{subcaption}
\usepackage{graphicx,subfig}
\usepackage{cleveref}
\usepackage{bxcoloremoji}

\usepackage{float}

\usepackage[utf8]{inputenc} %
\usepackage[T1]{fontenc}    %
\usepackage{hyperref}       %
\usepackage{url}            %
\usepackage{booktabs}       %
\usepackage{amsfonts}       %
\usepackage{nicefrac}       %
\usepackage{microtype}      %
\usepackage{graphicx}
\usepackage{subcaption} 
\usepackage{amssymb}
\usepackage{wrapfig}
\usepackage{lipsum}
\usepackage{enumitem}
\usepackage{stackengine}
\usepackage[font=small,labelfont=bf]{caption}
\usepackage{color}
\usepackage{adjustbox}
\usepackage{ulem}
\usepackage{rotating}
\usepackage{makecell}

\newcommand{\x}{\mathbf{x}}

\definecolor{blanchedalmond}{rgb}{1.0, 0.92, 0.8}
\definecolor{carmine}{rgb}{0.59, 0.0, 0.09}
\definecolor{lightblue}{rgb}{0.22,0.45,0.70}%

\renewcommand{\mathbf}{\boldsymbol}

\makeatletter
\def\Ddots{\mathinner{\mkern1mu\raise\p@
\vbox{\kern7\p@\hbox{.}}\mkern2mu
\raise4\p@\hbox{.}\mkern2mu\raise7\p@\hbox{.}\mkern1mu}}
\makeatother

\definecolor{amaranth}{rgb}{0.9, 0.17, 0.31}
\definecolor{antiquebrass}{rgb}{0.8, 0.58, 0.46}
\definecolor{antiquefuchsia}{rgb}{0.57, 0.36, 0.51}
\definecolor{chromeyellow}{rgb}{0.31, 0.47, 0.26}

\newcommand{\github}{\raisebox{-1.5pt}{\includegraphics[height=1.05em]{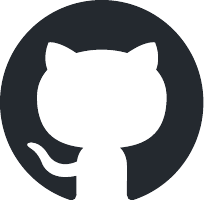}}}

\newtcolorbox{AIbox}[2][]{aibox,title=#2,#1}
\definecolor{lightblue}{rgb}{0.22,0.45,0.70}%
\definecolor{Gray}{gray}{0.95}
\definecolor{Cornsilk}{rgb}{1.0, 0.97, 0.86}

\definecolor{red_down_arrow}{HTML}{F04158}

\usepackage{amsmath}

\usepackage[all]{hypcap}

\title{Compositional Adversarial Training for Robust Visual Watermarking}

\runningtitle{Compositional Adversarial Training for Robust Visual Watermarking}

\author{
  Anirudh Satheesh$^1$,
  Michael-Andrei Panaitescu-Liess$^1$,
  Andrew Xu$^1$,
  Georgios Milis$^1$,
  Heng Huang$^1$,
  Zikui Cai$^1$, and
  Furong Huang
}

\affil[1]{University of Maryland}

\correspondingauthor{Anirudh Satheesh \href{https://anirudhsatheesh.com}{anirudhsatheesh.com}; Email \href{mailto:anirudhs@terpmail.umd.edu}{anirudhs@terpmail.umd.edu}}

\begin{document}

\begin{abstract}
Robust watermarking is typically trained with random post-processing augmentation, but random sampling under-covers the combinatorial space of realistic attack pipelines and rarely encounters the rare compositions that actually break detection. This leads to unstable training and poor sample efficiency. We instead formulate watermark robustness as a min-max problem over a structured space of compositional transformations. We propose Compositional Adversarial Training (CAT), a plug-in framework that learns a sequential differentiable adversary that observes the current watermarked image and selects an attack family at each step to maximally disrupt message recovery. CAT combines a straight-through Gumbel-Softmax attack selection with entropy regularization, allowing the backward pass to be end-to-end differentiable and aggregate gradient information across attack families, yielding faster, smoother convergence without collapsing to a single attack mode. We evaluate CAT on post-generation watermarks VideoSeal 0.0, VideoSeal 1.0, and PixelSeal and in-generation WMAR under both single-step and two-step attack suites, on in-distribution and multiple out-of-distribution image and video benchmarks. CAT consistently outperforms random-augmentation baselines trained with the same augmentation budget, with the largest gains on hard composed attacks and OOD evaluations; improving overall watermark capacity by up to 63.5\% in the single-step attack setting and 13.0\% in the compositional setting. In the autoregressive setting, CAT improves the TPR@FPR$=1\%$ by 12\% on average on difficult geometric transformations. These results show that robust visual watermarking benefits from training against adaptive compositional adversaries rather than independent random corruptions.

\coloremojicode{1F3E0} \textbf{Project Site}: \href{https://compositional-adversarial-training.github.io/}{https://compositional-adversarial-training.github.io/}

\github{} \textbf{Code Repository}: \href{https://github.com/Asatheesh6561/CAT}{https://github.com/Asatheesh6561/CAT}

\coloremojicode{1F917} \textbf{HuggingFace Models \& Datasets}: \href{https://huggingface.co/collections/asatheesh/cat}{https://huggingface.co/collections/asatheesh/cat}

\end{abstract}

\maketitle
\vspace{3mm}

\begin{figure}[!htbp]
    \centering
    \begin{subfigure}[t]{0.49\linewidth}
        \centering
        \includegraphics[width=\linewidth]{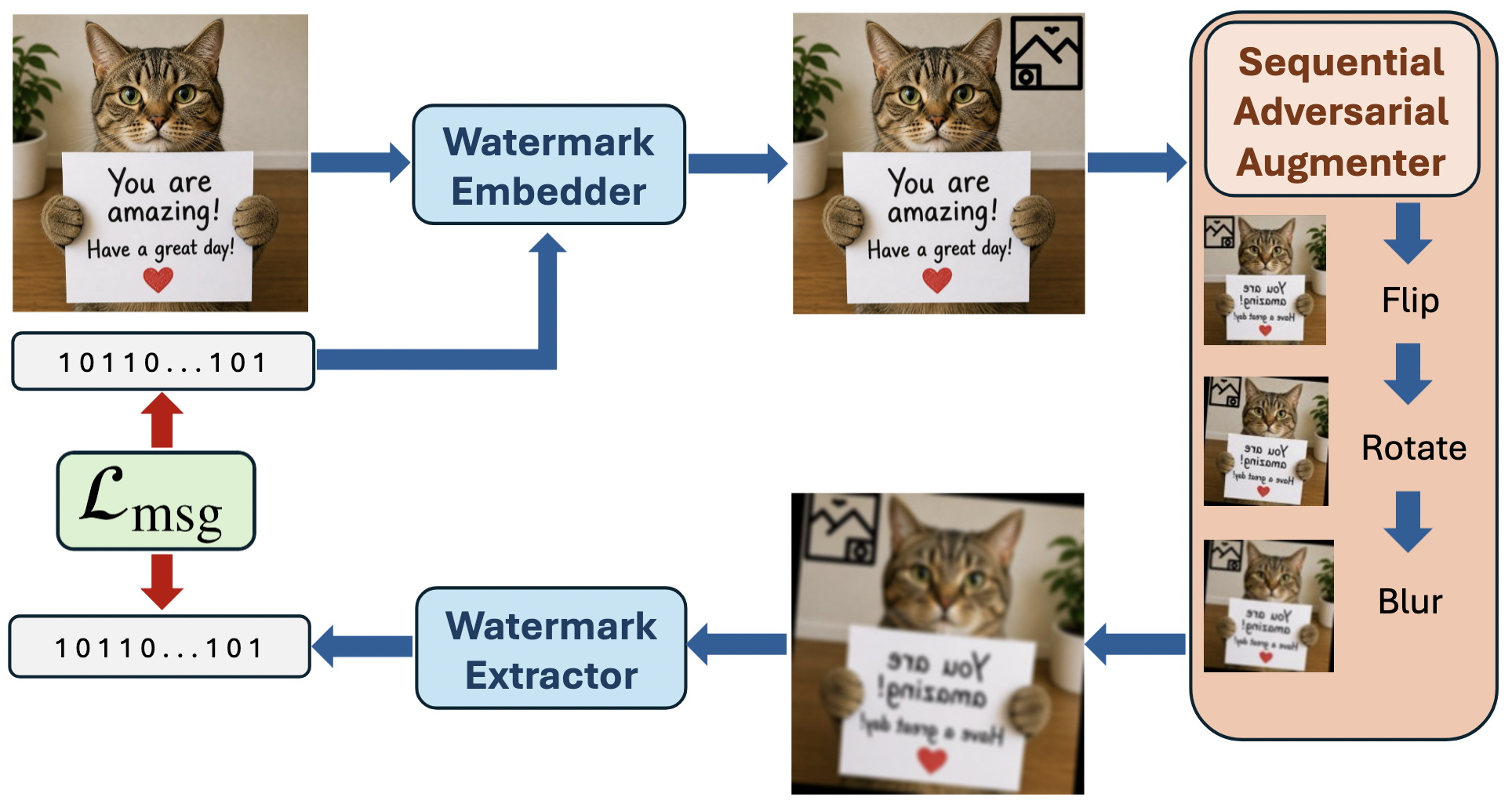}
        \label{fig:teaser_overview}
    \end{subfigure}\hfill
    \begin{subfigure}[t]{0.49\linewidth}
        \centering
        \includegraphics[width=\linewidth]{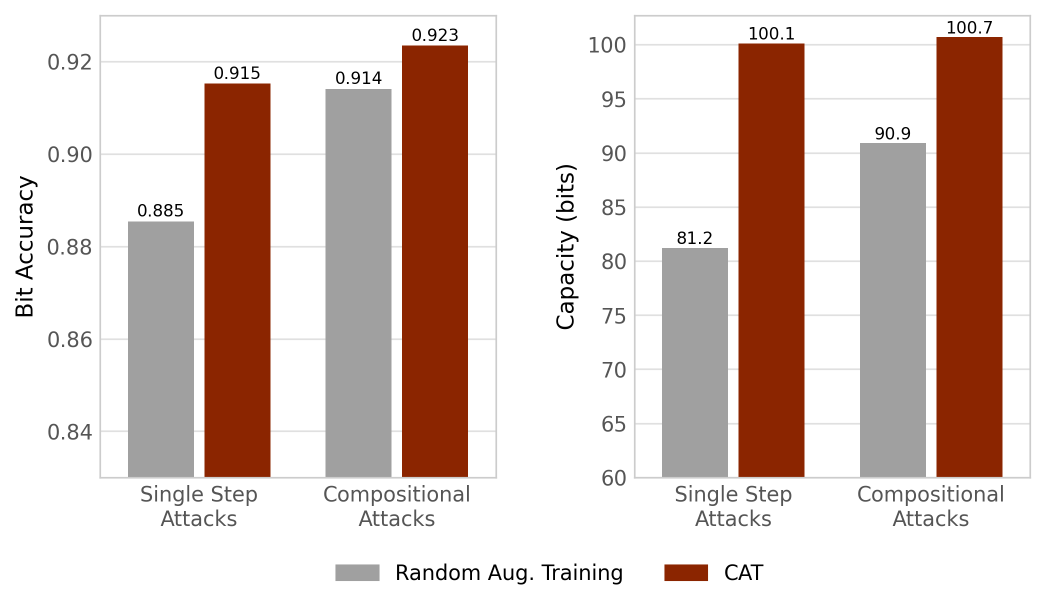}
        \label{fig:teaser_results}
    \end{subfigure}
    \caption{Overview of Compositional Adversarial Training (CAT) for visual watermarking. CAT improves the overall bit accuracy by 2.2\% and capacity by 17.0\% for single-step and compositional attacks.\label{fig:teaser}}
\end{figure}

\section{Introduction}
As synthetic and redistributed visual media proliferate, visual watermarking is increasingly used for provenance and misuse detection. In practice, watermarked images are rarely observed in their clean form and instead undergo recompression, resizing, cropping, blur, color shifts, and compositions of these operations \citep{jia2021mbrs, an2024waves, ding2024erasing}. Robustness therefore depends on surviving structured post-processing pipelines rather than isolated perturbations.

Current training relies primarily on random augmentation, which samples individual corruptions and parameters independently \citep{jovanovic2025watermarking, souvcek2025pixel}. While this improves performance under common distortions, it poorly reflects deployment conditions, where attacks are compositional, order-dependent, and highly imbalanced. As a result, training tends to overemphasize easy transformations while under-sampling rare but critical failure cases, leading to apparent robustness under standard evaluation yet brittle behavior under adaptive attack sequences.

\begin{figure}[t]
    \centering
    \captionsetup[subfigure]{justification=centering,singlelinecheck=false}
    \begin{subfigure}[t]{0.49\linewidth}
        \centering
        \includegraphics[width=\linewidth]{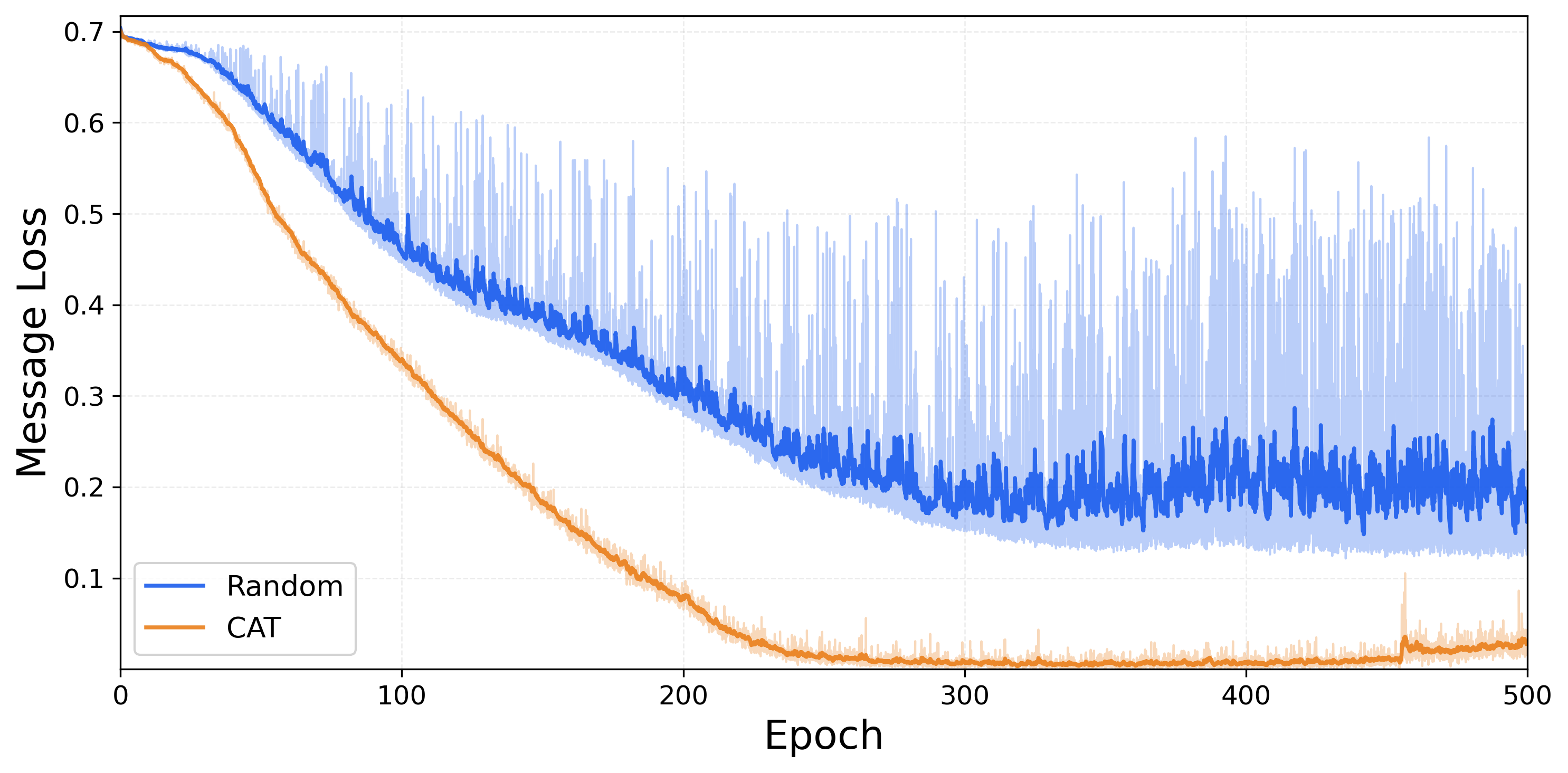}
        \caption{Single-step augmentation training.}
    \end{subfigure}\hfill
    \begin{subfigure}[t]{0.49\linewidth}
        \centering
        \includegraphics[width=\linewidth]{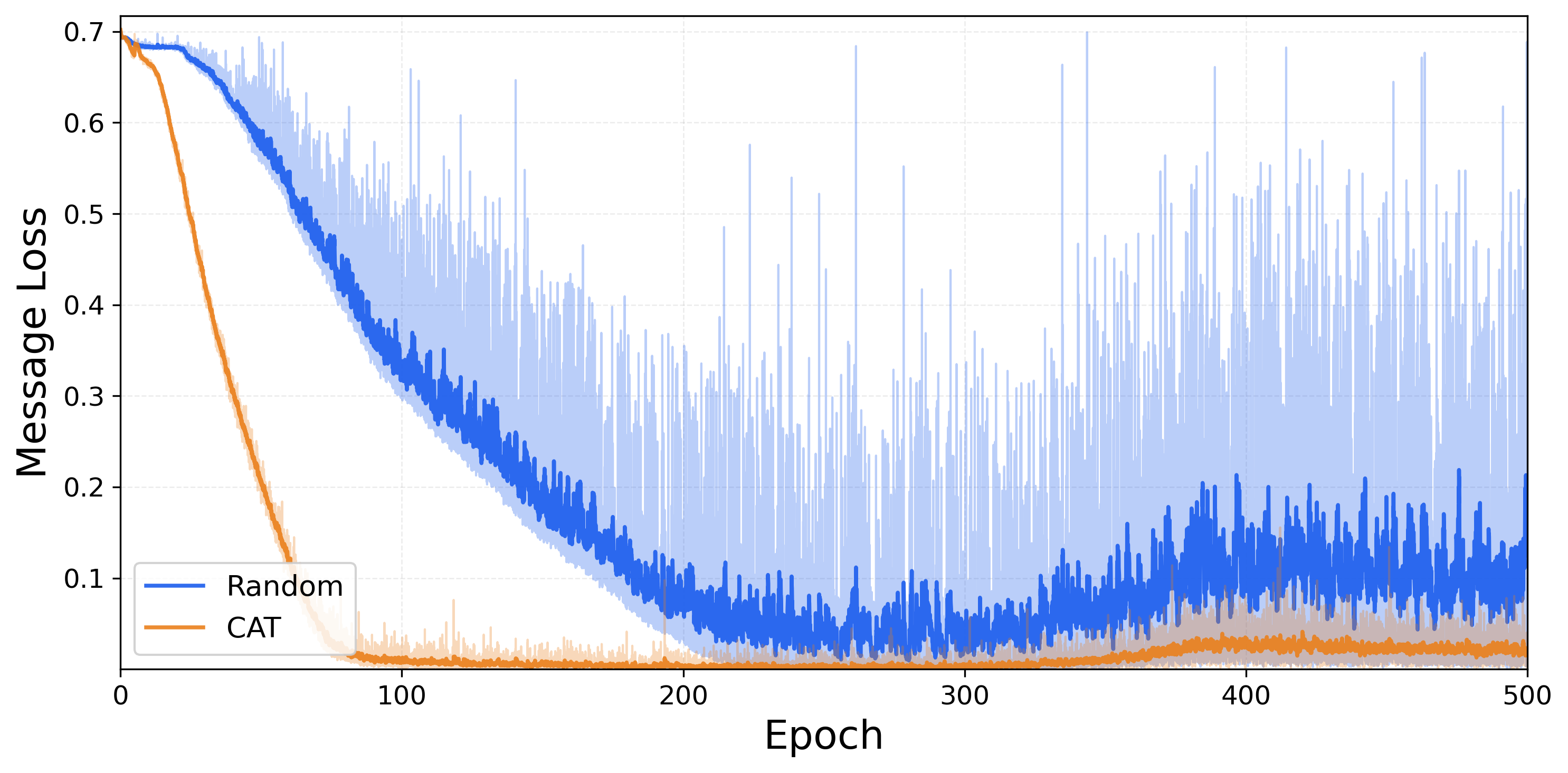}
        \caption{Compositional augmentation training.}
    \end{subfigure}
    \caption{\textbf{Random augmentation creates unstable training due to inefficient augmentation allocations, whereas the learned adversary consistently targets the model's current weaknesses.} Panel (a) shows the single-step setting and panel (b) the compositional setting. Under random augmentation, most sampled attacks are already easy and contribute little, while occasional hard or order-sensitive samples arrive sporadically, causing training to oscillate between uninformative updates and abrupt failures. CAT reduces this instability by adaptively concentrating on informative vulnerabilities, especially in the compositional regime where attack order matters.}
    \label{fig:rand_vs_adv}
\end{figure}

Pure adversarial training improves coverage but can be unstable in this setting, since an unconstrained inner maximization may rapidly converge to destructive transformations that remove the watermark entirely and collapse the learning signal \citep{wen2019romark}. In contrast, random augmentation is under-targeted and inefficient, as it fails to consistently surface the hard, order-sensitive failures that dominate real-world degradation, as shown in Figure~\ref{fig:rand_vs_adv}.

We propose Compositional Adversarial Training (CAT), which bridges these two regimes by restricting the adversary to a structured library of realistic, differentiable attack families while allowing adaptive composition over short sequences. A recurrent controller selects the attack type on the evolving image, enabling the discovery of order-sensitive attack chains such as blur followed by compression or resizing followed by color distortion. This process is made differentiable through a straight-through Gumbel-Softmax relaxation, while entropy regularization prevents premature collapse to a single dominant strategy.

Unlike inner-loop adversarial methods, CAT shares the main computational graph with the watermark model, preserves inference-time behavior, and introduces only 20--30\% additional training overhead. By explicitly optimizing over structured sequences of attacks, it consistently targets current model weaknesses rather than relying on randomly sampled perturbations. As illustrated in Figure~\ref{fig:rand_vs_adv}, this leads to more stable optimization and better coverage of difficult failure modes compared to random augmentation.

Empirically, CAT improves robustness across VideoSeal 0.0, VideoSeal 1.0 \citep{fernandez2024video}, and PixelSeal \citep{souvcek2025pixel} under both in-distribution and distribution-shifted evaluations. The gains are most pronounced under compositional attacks, where CAT achieves improvements of up to \textbf{63.5\%} under single-step attacks and \textbf{13.0\%} under composed attacks, while also converging faster than baseline methods. The same pattern extends to autoregressive watermarking: CAT substantially improves the hardest geometric failure modes, and CAT with synchronization reaches geometric TPR@FPR$=1\%$ of 0.71 on Taming and 0.72 on RAR-XL. Figure~\ref{fig:teaser} summarizes the setting and key results.

Our contributions are summarized as follows:
\begin{itemize}[leftmargin=1.5em]
  \item \textbf{Problem formulation.} We identify a core limitation of random augmentation for watermark robustness: it systematically under-emphasizes rare but difficult compositional attacks and over-emphasizes easy attacks, leading to brittle behavior and poor sample efficiency.

  \item \textbf{Method.} We introduce Compositional Adversarial Training (CAT), a learned sequential differentiable adversary that jointly selects attack identities and orders over short realistic attack compositions.

  \item \textbf{Optimization.} We show how to train CAT end-to-end using a straight-through Gumbel-Softmax controller, entropy regularization, and a shared single-pass update, enabling adaptive attack generation with weakness-aligned learning signals while avoiding policy collapse, keeping training practical, and leaving inference unchanged.

  \item \textbf{Empirical evidence.} We demonstrate consistent gains across multiple watermarking backbones, attack horizons, and out-of-distribution benchmarks, with the largest improvements on geometric and compositional attacks, alongside substantially faster convergence.
\end{itemize}

\section{Method}\label{sec:method}

\begin{figure}[t]
    \centering
    \includegraphics[width=0.8\linewidth]{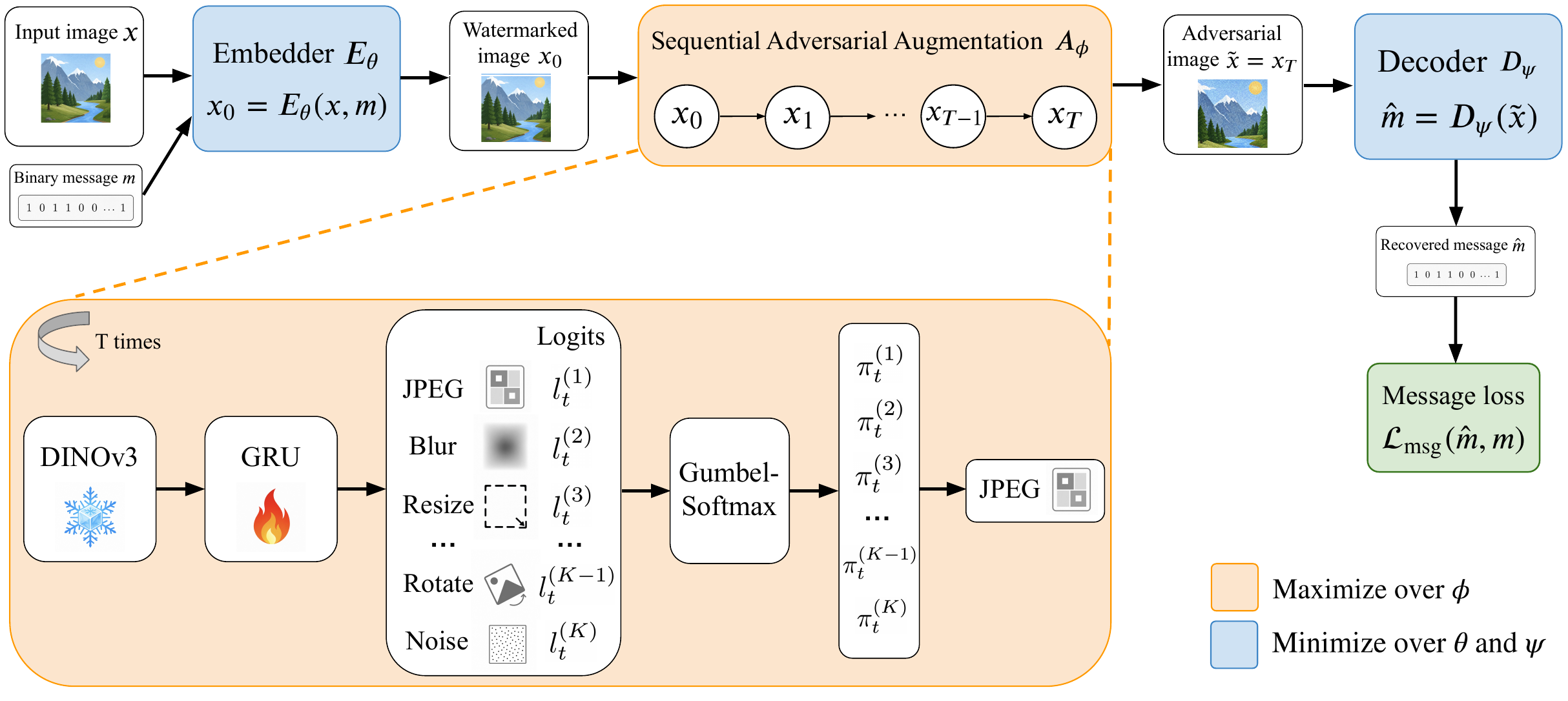}
    \caption{Conceptual overview of the proposed training pipeline. The embedder writes message $m$ into image $x$ to produce the watermarked image $x_0$. The adversary then repeatedly observes the current image $x_t$, uses a recurrent controller to produce logits, selects an attack family via straight-through Gumbel-Softmax, and applies differentiable attacks to obtain $x_{t+1}$. After $T$ steps, the final attacked image $x_T$ is passed to the extractor, and training optimizes message recovery loss. Entropy regularization, described in Section~\ref{sec:entropy}, keeps the attack policy diverse rather than collapsing to a single destructive sequence.} \label{fig:method_concept}
\end{figure}

\paragraph{Overview.}
Given a watermarked image, the adversary chooses a short sequence of attacks conditioned on the current intermediate image, relaxes discrete operation choice with a straight-through Gumbel-Softmax estimator, and regularizes the attack policy with entropy. The attack depth \(T\) controls the length of the composition: \(T=1\) corresponds to a single attack, while \(T=2\) enables two-step compositions such as blur followed by compression. In this work, we limit compositional augmentation to \(T=2\). While deeper compositions are in principle possible, we empirically find that extending to \(T \geq 3\) either requires substantially weakening individual augmentations or results in images that are severely degraded and no longer usable, making the watermarking task ill-posed. As a result, \(T=2\) provides a practical balance between compositional richness and perceptual fidelity. Because the adversary is differentiable and short-horizon, the inner problem can be optimized in the same training pass without an expensive search over long attack sequences. Figure~\ref{fig:method_concept} gives a conceptual overview of the full training pipeline.

\subsection{Problem Formulation and Sequential Adversarial Augmentation}\label{sec:formulation}

The goal of robust visual watermarking is to embed a hidden message into an image while preserving visual fidelity and enabling reliable decoding after post-processing. Let \(x \in \mathbb{R}^{H \times W \times C}\) denote a clean image and \(m \in \{0,1\}^{d_m}\) a binary message of length \(d_m\). The watermarking system consists of an embedder \(E_{\theta}\) and an extractor \(D_{\psi}\). The embedder produces a watermarked image \(x_{\mathrm{wm}} = E_{\theta}(x,m)\), which must satisfy an imperceptibility constraint, typically enforced via a perceptual similarity metric \(\mathcal{L}_{\mathrm{perc}}(x, x_{\mathrm{wm}}) \le \epsilon\). After embedding, the image is subjected to a post-processing attack \(t \in \mathcal{T}\), where \(\mathcal{T}\) is a bounded but expressive set of realistic transformations. Crucially, \(\mathcal{T}\) is compositional: an attack may consist of multiple parameterized operations applied in sequence, such as resizing, blurring, and compression in different orders. The extractor receives the attacked image \(\tilde{x} = t(x_{\mathrm{wm}})\) and predicts the recovered message \(\hat{m} = D_{\psi}(\tilde{x})\). We therefore seek parameters \((\theta, \psi)\) that minimize message recovery error under worst-case attacks from \(\mathcal{T}\). This yields the min--max objective
\begin{equation}
 \min_{\theta,\psi} \max_{\phi} \;
\mathbb{E}_{x,m} \Bigl[
\mathcal{L}_{\mathrm{msg}}\bigl(D_{\psi}(\mathcal{A}_{\phi}(E_{\theta}(x,m))),\, m\bigr)
\Bigr],
\label{eq:minmax}
\end{equation}
where \(\mathcal{A}_{\phi}\) is a learned adversary that parameterizes the compositional attack space \(\mathcal{T}\). The inner maximization ranges over attack identities, order, and continuous parameters, subject to bounded realistic ranges. Unlike conventional adversarial training based on small \(\ell_p\)-bounded perturbations, this objective captures the structured, order-sensitive nature of real-world image corruptions \citep{an2024waves, ding2024erasing}. It encourages robustness to realistic post-processing pipelines rather than to isolated perturbations. In practice, we evaluate robustness using two primary metrics: (i) bit accuracy, the fraction of correctly recovered bits, and (ii) usable capacity, the number of bits that can be reliably transmitted under attack.

\paragraph{Sequential attack generator.}
We instantiate \(\mathcal{A}_{\phi}\) as a controller over a library of \(K\) differentiable attack primitives \(\{f_i\}_{i=1}^K\). Starting from the watermarked image \(x_0 = E_{\theta}(x, m)\), the adversary applies
\begin{equation}
    x_{t+1} = f_{a_t}(x_t,\, \lambda_t), \quad t = 0, \ldots, T-1,
    \label{eq:sequential}
\end{equation}
where $a_t \in [K] := \{1,\ldots,K\}$ selects the attack primitive and $\lambda_t \in \Lambda_{a_t}$ denotes its continuous parameters, with each $\Lambda_i$ defining a bounded valid range for attack family $i$. The primitive library is built from standard differentiable post-processing operations used in robust watermarking, covering major corruption families---including value distortions, compression artifacts, and geometric transformations---while remaining visually plausible and efficiently composable during training. The final attacked image is $\tilde{x} = x_T$. Unlike a one-shot augmentation layer, this sequential generator lets later operations depend on the intermediate image produced by earlier ones, which is essential for order-sensitive compositions. In particular, depth $T=1$ recovers a single learned attack, while depth $T=2$ enables two-step attack compositions.

\paragraph{Differentiable attack selection.}
Optimizing \eqref{eq:minmax} requires gradients through the choice of $a_t$. Directly sampling a discrete operation would block backpropagation, and score-function estimators are high-variance because the loss depends on an entire composed attack sequence. We therefore use a straight-through Gumbel-Softmax relaxation \citep{jang2017categorical}. At each step $t$, a controller produces logits $\ell_t \in \mathbb{R}^K$ and samples
\begin{equation}
    \pi_t = \operatorname{GS}(\ell_t,\, \tau) \in \Delta^{K-1},
    \label{eq:gumbel}
\end{equation}
where $\tau$ is the temperature. The forward pass uses a one-hot sample, while the backward pass uses the continuous relaxation so that gradients can flow through attack choice. All $K$ primitives are evaluated in parallel with parameters $\lambda_t^{(i)} \sim \mathcal{U}(\Lambda_i)$, where $\mathcal{U}$ denotes the uniform distribution over the valid range $\Lambda_i$, and their outputs are combined as
\begin{equation}
    x_{t+1} = \sum_{i=1}^{K} \pi_t^{(i)}\, f_i(x_t,\, \lambda_t^{(i)}).
    \label{eq:mixture}
\end{equation}
\paragraph{Relaxed inner maximization.}
Combining \eqref{eq:sequential} and \eqref{eq:mixture}, the original min-max problem becomes
\begin{equation}
    \min_{\theta,\psi} \max_{\phi} \;
    \mathbb{E}_{x,m,\, a \sim \pi_\phi}\Bigl[
        \mathcal{L}_{\mathrm{msg}}\bigl(D_\psi(x_T),\, m\bigr)
    \Bigr],
    \label{eq:relaxed_minmax}
\end{equation}
where $x_T$ is the attacked image after $T$ sequential steps and $a = (a_0, \ldots, a_{T-1})$ denotes the full attack sequence. Equivalently, for fixed watermark parameters $(\theta,\psi)$, the adversary solves
\begin{equation}
    \max_{\phi} \; \mathbb{E}_{a \sim \pi_\phi}\bigl[\mathcal{L}_{\mathrm{msg}}\bigl(D_\psi(x_T),\, m\bigr)\bigr],
    \label{eq:inner_max_sequence}
\end{equation}
This highlights the main novelty of our setting: the adversary must learn not only which corruption is harmful, but also which order of corruptions jointly maximize decoder failure. Because attack choice is relaxed and differentiable, the controller can be updated jointly with the watermark model rather than through a separate black-box search over attack sequences.

\subsection{Entropy Regularization}\label{sec:entropy}

Without additional constraints, the inner maximization in Eq.~\eqref{eq:inner_max_sequence} can collapse to a single maximally damaging attack sequence,
\begin{equation}
    a^\star = \arg\max_a \; \mathcal{L}_{\mathrm{msg}}\bigl(D_\psi(x_T),\, m\bigr),
\end{equation}
which causes the adversary to over-specialize and reduces exploration as training evolves. To prevent this, we add an entropy bonus to the attack policy and optimize
\begin{equation}
    \max_{\phi} \;
    \mathbb{E}_{a \sim \pi_\phi}\bigl[\mathcal{L}_{\mathrm{msg}}\bigl(D_\psi(x_T),\, m\bigr)\bigr]
    + \lambda_{\text{ent}}\, \mathcal{H}(\pi_\phi),
    \label{eq:entropy_obj}
\end{equation}
where $\mathcal{H}(\pi_\phi)$ is the Shannon entropy of the attack policy and $\lambda_{\text{ent}}$ controls the exploration--strength tradeoff. In practice, this keeps the controller exploratory early in training, avoids premature collapse to a trivially destructive strategy, and improves robustness beyond a single dominant failure mode. Full implementation details, including the per-step entropy decomposition and the exact optimization objective, are provided in Appendix~\ref{sec:further-implementation-details}.

\section{Related Work}

\paragraph{Deep Learning-Based Watermarking} A more comprehensive review appears in Appendix~\ref{sec:extended-related-work}. Deep neural networks have become the dominant approach to image and video watermarking, largely following the encoder--noise--decoder paradigm introduced by HiDDeN~\citep{zhu2018hidden}. In this framework, an encoder embeds a message into an image, a noise layer simulates distortions, and a decoder recovers the message, enabling end-to-end optimization of robustness and imperceptibility. This formulation has been widely applied to images in MBRS~\citep{jia2021mbrs}, TrustMark~\citep{bui2025trustmark}, and InvisMark~\citep{xu2025invismark}, and to video through DVMark~\citep{luo2023dvmark}, VideoSeal~\citep{fernandez2024video}, and PixelSeal~\citep{souvcek2025pixel}. A further line of work extends the same paradigm to in-generation watermarking methods such as Stable Signature~\citep{fernandez2023stable}, Tree-Ring~\citep{wen2023tree}, and WMAR~\citep{jovanovic2025watermarking}. Despite these architectural advances, robustness is still typically trained through fixed or randomly sampled augmentations. That strategy helps against common isolated corruptions, but it under-covers realistic attack pipelines in which operation identity, parameter range, and order all matter. Our work addresses this training gap by replacing blind random augmentation with a learned compositional adversary that adaptively targets current failure modes.

\textbf{Adversarial Training for Watermarking.}
While adversarial training has been explored in watermarking, existing approaches remain limited in scope. Many works introduce adversarial components but target objectives orthogonal to post-processing robustness. For example, HiDDeN~\citep{zhu2018hidden} and PixelSeal~\citep{souvcek2025pixel} employ discriminators to enforce visual imperceptibility rather than robustness; ROBIN~\citep{huang2024robin} focuses on watermark concealment during diffusion generation; and InvZW~\citep{tanvir2025invzw} applies adversarial learning to zero-watermarking, which avoids modifying the image and therefore bypasses the encoder--extractor robustness tradeoff entirely. Methods that explicitly target robustness, such as RoMark~\citep{wen2019romark}, formulate watermarking as a min--max optimization problem but are not end-to-end differentiable and require repeated decoder evaluations to approximate worst-case augmentations, which makes training expensive and still leaves sequential compositions largely unmodeled. In contrast, CAT introduces a fully differentiable adversary over a structured library of composable transformations. It jointly optimizes attack identity, order, and continuous parameters within the same training graph, allowing the model to train against hard short attack sequences that fixed or randomly sampled augmentation schemes systematically miss.

\section{Experiments}
\label{sec:experiments}

\subsection{Architecture and Implementation Details}
\label{sec:arch_train}

\paragraph{Architecture and training details.}
We evaluate CAT on VideoSeal 0.0, VideoSeal 1.0 \citep{fernandez2024video}, and PixelSeal \citep{souvcek2025pixel}. The underlying embedder and extractor follow the original PixelSeal-style architectures, while the learned adversary uses frozen DINOv3 ViT-S/16 features, a lightweight GRU for attack selection, and a lightweight MLP attack head to select augmentations sequentially. In total, the adversary contributes only 800K trainable parameters and modest memory overhead. Full architecture and training hyperparameters are deferred to Appendix~\ref{sec:further-implementation-details}.

At each adversarial step $t$, visual features $z_t \in \mathbb{R}^{d_h}$ are extracted from $x_t$ using a frozen DINOv3 ViT-S/16 backbone \citep{simeoni2025dinov3} followed by a two-layer MLP projection, and a GRU updates
\begin{equation}
    h_t = \operatorname{GRU}(z_t,\, h_{t-1}), \quad h_0 = \mathbf{0},
    \label{eq:gru}
\end{equation}
so that attack choices can depend on previously applied transformations. This recurrent controller is the core architectural ingredient needed to study sequential, order-sensitive attacks; implementation specifics are given in Appendix~\ref{sec:further-implementation-details}.

\paragraph{Evaluation protocol.}
For each learned backbone, we compare a compute-matched baseline trained with the same augmentation budget but without the learned adversary, and the corresponding model trained with Compositional Adversarial Training (CAT) under a set of predefined augmentations. We also compare to the pretrained variants of MBRS \citep{jia2021mbrs}, TrustMark \citep{bui2025trustmark}, and InvisMark \citep{xu2025invismark}. The single-step attack setting corresponds to $T=1$, while the compositional attack setting corresponds to $T=2$ and therefore a two-step attack sequence. Unless otherwise noted, the tables report bit accuracy and watermark capacity for each attack family, along with their overall averages; higher is better for both metrics. In the main image experiments, we evaluate on 1000 in-distribution (ID) SA-1B test images \citep{kirillov2023segment} and on the out-of-distribution (OOD) CLIC benchmark \citep{aytekin2019compression}. This compute-matched comparison ensures that any gains from CAT are not simply due to additional augmentation budget, but to adaptive adversarial training itself.

\subsection{Robustness results (Single-step attacks)}

\paragraph{CAT is most effective on challenging attack settings.}
Table~\ref{tab:results_depth1_id_ood} shows that even a single learned attack step improves robustness under a fair compute-matched comparison. On SA-1B, CAT improves overall capacity for all three learned backbones, from 81.06 to 82.85 for VideoSeal 0.0, from 106.41 to 125.57 for VideoSeal 1.0, and from 56.21 to 91.91 for PixelSeal. The same pattern holds on CLIC, where overall capacity rises from 80.25 to 81.84 for VideoSeal 0.0, from 105.00 to 123.00 for VideoSeal 1.0, and from 55.26 to 89.44 for PixelSeal. The gains are concentrated in the more difficult corruption families, especially geometric and combined attacks, where random augmentation is least likely to sample the failure-inducing cases that matter most.

\paragraph{The primary empirical benefit is increased usable capacity.}
CAT yields not only improvements in bit accuracy, but also substantial gains in usable watermark capacity. This effect is especially pronounced for PixelSeal, whose overall capacity rises by 35.70 points on SA-1B and by 34.18 points on CLIC, and for VideoSeal 1.0, whose overall capacity improves by 19.16 points on SA-1B and 18.00 points on CLIC. Here, capacity measures the effective number of reliably recoverable bits under attack, not just whether decoding succeeds on average. This distinction matters because capacity increases sharply as bit accuracy approaches 100\%, so even modest average accuracy gains can correspond to much larger capacity gains once performance enters the high-accuracy regime. The table also shows that CAT-trained models remain competitive with or stronger than pretrained baselines on the main image benchmark: for example, PixelSeal with CAT reaches 91.91 overall capacity on SA-1B, exceeding InvisMark (71.01) and TrustMark (76.08), although MBRS remains higher because it operates with a larger payload.

\begin{table}[htbp]
\caption{\textbf{A single learned attack step already improves robustness over random augmentation training, with the largest gains for VideoSeal 1.0 and PixelSeal, especially on the most difficult combined attacks and under severe distribution shift.} Single-step attack accuracy and watermark capacity results on in-distribution SA-1B and out-of-distribution CLIC images. Each cell reports bit accuracy and capacity for identity, value, compression, geometric, and combined augmentation families. For VideoSeal and PixelSeal, the `+ CAT` rows use Compositional Adversarial Training, while the corresponding base rows use random augmentation training; higher is better for both metrics.}
\centering
\resizebox{\textwidth}{!}{
\begin{tabular}{l l*{6}{cc}}
\toprule
 &  & \multicolumn{2}{c}{Identity} & \multicolumn{2}{c}{Value} & \multicolumn{2}{c}{Compression} & \multicolumn{2}{c}{Geometric} & \multicolumn{2}{c}{Combined} & \multicolumn{2}{c}{Overall} \\
\cmidrule(lr){3-4}\cmidrule(lr){5-6}\cmidrule(lr){7-8}\cmidrule(lr){9-10}\cmidrule(lr){11-12}\cmidrule(lr){13-14}
 & Model (bits) & Bit acc. $(\uparrow)$ & Capacity $(\uparrow)$ & Bit acc. $(\uparrow)$ & Capacity $(\uparrow)$ & Bit acc. $(\uparrow)$ & Capacity $(\uparrow)$ & Bit acc. $(\uparrow)$ & Capacity $(\uparrow)$ & Bit acc. $(\uparrow)$ & Capacity $(\uparrow)$ & Bit acc. $(\uparrow)$ & Capacity $(\uparrow)$ \\
\midrule
\multirow{9}{*}{\rotatebox{90}{\textbf{SA-1B (ID)}}} & InvisMark (100) & 0.990 & 95.99 & 0.876 & 72.90 & 0.954 & 88.45 & 0.828 & 63.25 & 0.861 & 67.84 & 0.869 & 71.01 \\
 & TrustMark (100) & 0.996 & 98.92 & 0.956 & 86.28 & 0.898 & 79.09 & 0.754 & 50.29 & 0.993 & 97.90 & 0.889 & 76.08 \\
 & MBRS (256) & 0.987 & 242.60 & 0.915 & 185.99 & 0.884 & 190.52 & 0.653 & 78.90 & 0.959 & 217.48 & 0.834 & 158.92 \\
\cmidrule(lr){2-14}
\morecmidrules\cmidrule(lr){2-14}
 & VideoSeal 0.0 (96) & 0.997 & 94.00 & 0.984 & 88.69 & 0.980 & 86.86 & 0.945 & 75.58 & 0.994 & 92.33 & 0.961 & 81.06 \\
 & \quad + CAT & 0.998 & 94.75 & 0.978 & 86.93 & 0.986 & 89.21 & 0.953 & 78.60 & 0.994 & 92.22 & \textbf{0.966}$_{\textcolor{ForestGreen}{\footnotesize \uparrow 0.5\%}}$ & \textbf{82.85}$_{\textcolor{ForestGreen}{\footnotesize \uparrow 2.2\%}}$ \\
\cmidrule(lr){2-14}
 & VideoSeal 1.0 (256) & 0.898 & 135.17 & 0.879 & 123.45 & 0.872 & 118.96 & 0.822 & 94.39 & 0.892 & 130.60 & 0.846 & 106.41 \\
 & \quad + CAT & 0.941 & 175.63 & 0.857 & 129.47 & 0.896 & 146.11 & 0.835 & 114.75 & 0.921 & 160.39 & \textbf{0.854}$_{\textcolor{ForestGreen}{\footnotesize \uparrow 1.0\%}}$ & \textbf{125.57}$_{\textcolor{ForestGreen}{\footnotesize \uparrow 18.0\%}}$ \\
\cmidrule(lr){2-14}
 & PixelSeal (128) & 0.918 & 76.46 & 0.895 & 68.65 & 0.880 & 63.89 & 0.819 & 48.19 & 0.902 & 70.27 & 0.849 & 56.21 \\
 & \quad + CAT & 0.986 & 117.73 & 0.956 & 102.90 & 0.957 & 102.96 & 0.900 & 83.05 & 0.973 & 109.36 & \textbf{0.925}$_{\textcolor{ForestGreen}{\footnotesize \uparrow 9.0\%}}$ & \textbf{91.91}$_{\textcolor{ForestGreen}{\footnotesize \uparrow 63.5\%}}$ \\
\midrule
\multirow{9}{*}{\rotatebox{90}{\textbf{CLIC (OOD)}}} & InvisMark (100) & 0.990 & 96.02 & 0.876 & 72.93 & 0.691 & 35.10 & 0.825 & 62.83 & 0.582 & 19.23 & 0.764 & 51.59 \\
 & TrustMark (100) & 0.993 & 98.39 & 0.959 & 87.60 & 0.894 & 78.18 & 0.753 & 50.21 & 0.947 & 84.07 & 0.878 & 73.03 \\
 & MBRS (256) & 0.987 & 243.37 & 0.922 & 191.47 & 0.848 & 162.21 & 0.648 & 74.77 & 0.775 & 105.06 & 0.786 & 128.44 \\
\cmidrule(lr){2-14}
\morecmidrules\cmidrule(lr){2-14}
 & VideoSeal 0.0 (96) & 0.998 & 94.59 & 0.986 & 89.85 & 0.977 & 86.11 & 0.948 & 76.61 & 0.844 & 57.95 & 0.956 & 80.25 \\
 & \quad + CAT & 0.999 & 95.18 & 0.974 & 86.34 & 0.982 & 87.96 & 0.957 & 79.74 & 0.860 & 61.83 & \textbf{0.961}$_{\textcolor{ForestGreen}{\footnotesize \uparrow 0.4\%}}$ & \textbf{81.84}$_{\textcolor{ForestGreen}{\footnotesize \uparrow 2.0\%}}$ \\
\cmidrule(lr){2-14}
 & VideoSeal 1.0 (256) & 0.900 & 136.47 & 0.883 & 125.61 & 0.869 & 117.38 & 0.824 & 95.59 & 0.766 & 78.96 & 0.842 & 105.00 \\
 & \quad + CAT & 0.946 & 179.81 & 0.853 & 129.33 & 0.884 & 140.37 & 0.840 & 118.13 & 0.723 & 81.29 & \textbf{0.846}$_{\textcolor{ForestGreen}{\footnotesize \uparrow 0.5\%}}$ & \textbf{123.00}$_{\textcolor{ForestGreen}{\footnotesize \uparrow 17.1\%}}$ \\
\cmidrule(lr){2-14}
 & PixelSeal (128) & 0.921 & 77.60 & 0.900 & 70.26 & 0.871 & 61.44 & 0.822 & 49.03 & 0.748 & 40.15 & 0.844 & 55.26 \\
 & \quad + CAT & 0.992 & 121.32 & 0.946 & 101.38 & 0.948 & 99.94 & 0.899 & 83.02 & 0.842 & 72.27 & \textbf{0.916}$_{\textcolor{ForestGreen}{\footnotesize \uparrow 8.5\%}}$ & \textbf{89.44}$_{\textcolor{ForestGreen}{\footnotesize \uparrow 61.8\%}}$ \\
\bottomrule
\end{tabular}
}
\label{tab:results_depth1_id_ood}
\end{table}

\subsection{Robustness results (Compositional attacks)}

\paragraph{CAT is particularly beneficial for compositional attacks.}
The advantage of CAT becomes clearer in the compositional setting, where the adversary applies a two-step augmentation sequence ($T=2$) and must model both attack identity and attack order. Table~\ref{tab:results_d2_id_ood} shows clear overall capacity gains for every learned backbone on both datasets: on SA-1B, overall capacity rises from 77.79 to 86.03 for VideoSeal 0.0, from 96.13 to 108.30 for VideoSeal 1.0, and from 98.76 to 107.73 for PixelSeal; on CLIC, it rises from 77.31 to 87.34, from 96.97 to 109.24, and from 96.20 to 106.43, respectively. These gains are concentrated on the harder mixed and repeated attack pairs. For example, on SA-1B, VideoSeal 1.0 improves from 134.39 to 172.97 on Comp+Comp and from 109.22 to 141.66 on Comp+Geom, while PixelSeal improves from 117.75 to 126.29 on Comp+Comp and from 106.43 to 116.19 on Comp+Geom. This pattern is consistent with CAT learning order-sensitive robustness rather than merely increasing exposure to more transformations.

\paragraph{The strongest compositional robustness gains appear in usable capacity and even surpass pretrained baselines with more bits.}
As in the single-step setting, the principal benefit of CAT under compositional attacks is a substantial increase in usable watermark capacity. The largest overall gains appear for VideoSeal 1.0 and PixelSeal: on SA-1B, VideoSeal 1.0 improves by 12.17 capacity points and PixelSeal by 8.97, while on CLIC the corresponding gains are 12.27 and 10.23. As above, capacity is the most informative metric near the high-accuracy regime, so gains here indicate a more reliable margin of correct decoding across attack types rather than isolated wins on a narrow subset of perturbations. Relative to the pretrained baselines in the same table, CAT-trained PixelSeal reaches 107.73 overall capacity on SA-1B and 106.43 on CLIC, exceeding both InvisMark and TrustMark on each benchmark. Because compositional attacks amplify weaknesses that may not appear under single transformations, these gains suggest that CAT improves the structure of the learned robustness rather than tuning to a narrow benchmark.

\begin{table}[htbp]
\caption{\textbf{Under compositional attacks, CAT yields improvements in usable watermark capacity, especially for VideoSeal 1.0 and PixelSeal on hard mixed and repeated attack pairs.} Compositional-attack results on in-distribution SA-1B and out-of-distribution CLIC images, where the adversary applies a two-step attack sequence. Each cell reports bit accuracy and capacity for pairwise augmentation-family compositions; higher is better for both metrics. For VideoSeal and PixelSeal, the `+ CAT` rows use Compositional Adversarial Training, while the corresponding base rows use the compute-matched baseline without the learned adversary.}
\centering
\resizebox{\textwidth}{!}{
\begin{tabular}{l l*{7}{cc}}
\toprule
 &  & \multicolumn{2}{c}{Val+Val} & \multicolumn{2}{c}{Val+Comp} & \multicolumn{2}{c}{Val+Geom} & \multicolumn{2}{c}{Comp+Comp} & \multicolumn{2}{c}{Comp+Geom} & \multicolumn{2}{c}{Geom+Geom} & \multicolumn{2}{c}{Overall} \\
\cmidrule(lr){3-4}\cmidrule(lr){5-6}\cmidrule(lr){7-8}\cmidrule(lr){9-10}\cmidrule(lr){11-12}\cmidrule(lr){13-14}\cmidrule(lr){15-16}
 & Model (bits) & Bit acc. $(\uparrow)$ & Capacity $(\uparrow)$ & Bit acc. $(\uparrow)$ & Capacity $(\uparrow)$ & Bit acc. $(\uparrow)$ & Capacity $(\uparrow)$ & Bit acc. $(\uparrow)$ & Capacity $(\uparrow)$ & Bit acc. $(\uparrow)$ & Capacity $(\uparrow)$ & Bit acc. $(\uparrow)$ & Capacity $(\uparrow)$ & Bit acc. $(\uparrow)$ & Capacity $(\uparrow)$ \\
\midrule
\multirow{9}{*}{\rotatebox{90}{\textbf{SA-1B (ID)}}} & InvisMark (100) & 0.898 & 71.44 & 0.652 & 27.15 & 0.875 & 68.63 & 0.474 & 0.85 & 0.603 & 21.75 & 0.888 & 70.52 & 0.813 & 57.92 \\
 & TrustMark (100) & 0.957 & 86.21 & 0.965 & 88.49 & 0.786 & 52.86 & 0.974 & 90.92 & 0.779 & 50.97 & 0.708 & 36.58 & 0.834 & 62.14 \\
 & MBRS (256) & 0.917 & 182.87 & 0.836 & 130.67 & 0.602 & 41.86 & 0.774 & 90.86 & 0.554 & 15.81 & 0.495 & 0.98 & 0.653 & 60.74 \\
\cmidrule(lr){2-16}
\morecmidrules\cmidrule(lr){2-16}
 & VideoSeal 0.0 (96) & 0.972 & 83.44 & 0.985 & 87.84 & 0.961 & 78.08 & 0.992 & 91.40 & 0.974 & 82.79 & 0.930 & 72.83 & 0.961 & 77.79 \\
 & \quad + CAT & 0.988 & 90.30 & 0.995 & 93.28 & 0.979 & 86.26 & 0.999 & 95.34 & 0.990 & 90.79 & 0.935 & 78.05 & \textbf{0.978}$_{\textcolor{ForestGreen}{\footnotesize \uparrow 1.8\%}}$ & \textbf{86.03}$_{\textcolor{ForestGreen}{\footnotesize \uparrow 10.6\%}}$ \\
\cmidrule(lr){2-16}
 & VideoSeal 1.0 (256) & 0.875 & 121.35 & 0.887 & 127.69 & 0.827 & 95.41 & 0.897 & 134.39 & 0.858 & 109.22 & 0.799 & 86.78 & \textbf{0.829} & 96.13 \\
 & \quad + CAT & 0.840 & 121.57 & 0.891 & 145.41 & 0.826 & 109.69 & 0.938 & 172.97 & 0.894 & 141.66 & 0.825 & 111.74 & 0.827$_{\textcolor{red_down_arrow}{\footnotesize \downarrow 0.3\%}}$ & \textbf{108.30}$_{\textcolor{ForestGreen}{\footnotesize \uparrow 12.7\%}}$ \\
\cmidrule(lr){2-16}
 & PixelSeal (128) & 0.964 & 106.45 & 0.977 & 112.06 & 0.954 & 100.25 & 0.987 & 117.75 & 0.968 & 106.43 & 0.922 & 93.04 & 0.952 & 98.76 \\
 & \quad + CAT & 0.974 & 113.81 & 0.990 & 121.27 & 0.964 & 107.65 & 0.998 & 126.29 & 0.982 & 116.19 & 0.928 & 100.21 & \textbf{0.965}$_{\textcolor{ForestGreen}{\footnotesize \uparrow 1.4\%}}$ & \textbf{107.73}$_{\textcolor{ForestGreen}{\footnotesize \uparrow 9.1\%}}$ \\
\midrule
\multirow{9}{*}{\rotatebox{90}{\textbf{CLIC (OOD)}}} & InvisMark (100) & 0.902 & 72.69 & 0.653 & 27.61 & 0.875 & 68.97 & 0.474 & 0.85 & 0.600 & 21.29 & 0.883 & 69.50 & 0.812 & 57.96 \\
 & TrustMark (100) & 0.964 & 88.76 & 0.968 & 89.85 & 0.790 & 53.95 & 0.974 & 91.65 & 0.782 & 51.93 & 0.710 & 37.13 & 0.838 & 63.39 \\
 & MBRS (256) & 0.927 & 190.84 & 0.851 & 140.04 & 0.608 & 45.61 & 0.791 & 100.83 & 0.559 & 18.03 & 0.496 & 0.97 & 0.660 & 65.18 \\
\cmidrule(lr){2-16}
\morecmidrules\cmidrule(lr){2-16}
 & VideoSeal 0.0 (96) & 0.978 & 86.28 & 0.961 & 81.15 & 0.968 & 80.81 & 0.937 & 77.05 & 0.947 & 75.52 & 0.937 & 74.67 & 0.957 & 77.31 \\
 & \quad + CAT & 0.992 & 91.87 & 0.991 & 91.84 & 0.984 & 88.57 & 0.977 & 89.05 & 0.988 & 89.97 & 0.940 & 79.73 & \textbf{0.981}$_{\textcolor{ForestGreen}{\footnotesize \uparrow 2.6\%}}$ & \textbf{87.34}$_{\textcolor{ForestGreen}{\footnotesize \uparrow 13.0\%}}$ \\
\cmidrule(lr){2-16}
 & VideoSeal 1.0 (256) & 0.881 & 124.44 & 0.886 & 127.38 & 0.832 & 98.03 & 0.895 & 133.09 & 0.854 & 106.99 & 0.800 & 87.42 & \textbf{0.831} & 96.97 \\
 & \quad + CAT & 0.852 & 128.33 & 0.869 & 136.19 & 0.839 & 117.17 & 0.909 & 157.25 & 0.878 & 134.59 & 0.831 & 115.09 & 0.827$_{\textcolor{red_down_arrow}{\footnotesize \downarrow 0.5\%}}$ & \textbf{109.24}$_{\textcolor{ForestGreen}{\footnotesize \uparrow 12.7\%}}$ \\
\cmidrule(lr){2-16}
 & PixelSeal (128) & 0.973 & 111.00 & 0.939 & 98.69 & 0.961 & 103.61 & 0.902 & 90.40 & 0.926 & 92.72 & 0.927 & 94.93 & 0.941 & 96.20 \\
 & \quad + CAT & 0.977 & 115.17 & 0.980 & 115.86 & 0.967 & 108.95 & 0.969 & 114.39 & 0.970 & 110.31 & 0.929 & 100.63 & \textbf{0.962}$_{\textcolor{ForestGreen}{\footnotesize \uparrow 2.3\%}}$ & \textbf{106.43}$_{\textcolor{ForestGreen}{\footnotesize \uparrow 10.6\%}}$ \\
\bottomrule
\end{tabular}
}
\label{tab:results_d2_id_ood}
\end{table}

\subsection{Validation Convergence}

\begin{figure*}[t]
    \centering
    \captionsetup[subfigure]{justification=centering,singlelinecheck=false}
    \begin{subfigure}{0.49\textwidth}
        \centering
        \includegraphics[width=\linewidth]{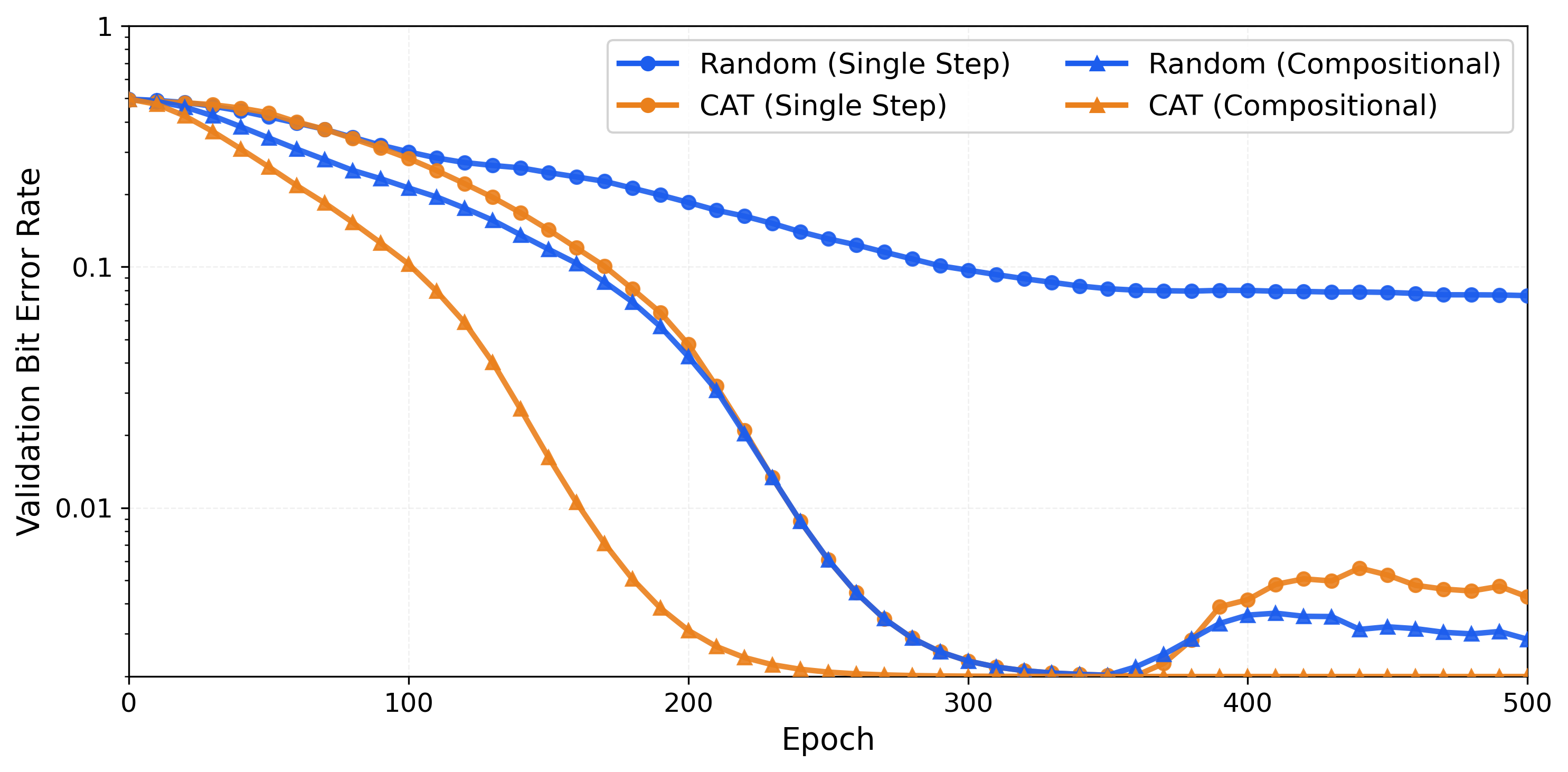}
        \caption{PixelSeal}
        \label{fig:pixelseal}
    \end{subfigure}
    \hfill
    \begin{subfigure}{0.49\textwidth}
        \centering
        \includegraphics[width=\linewidth]{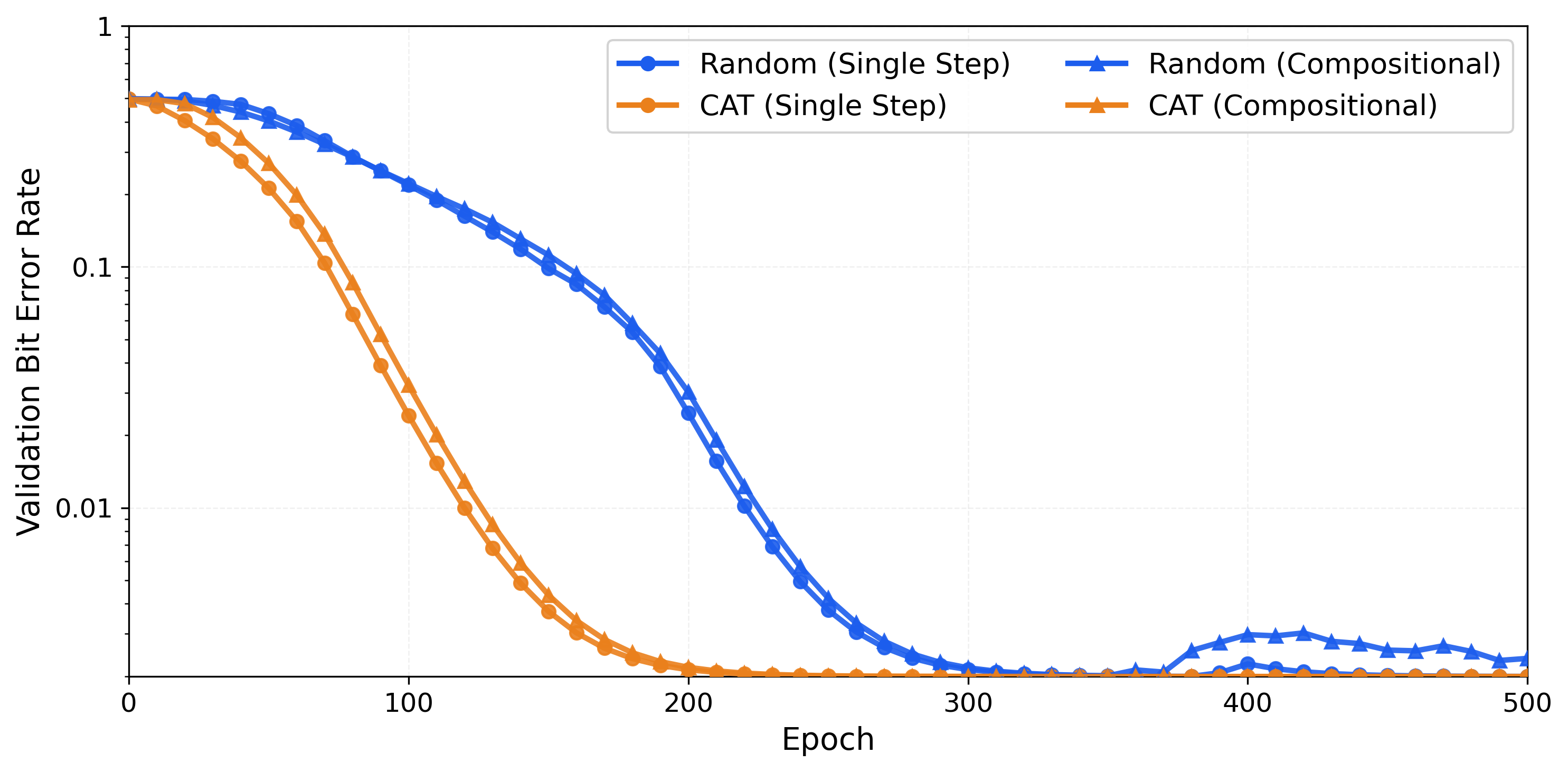}
        \caption{VideoSeal}
        \label{fig:videoseal}
    \end{subfigure}
    \caption{\textbf{CAT substantially accelerates convergence for both PixelSeal and VideoSeal, and this advantage persists from single-step to compositional training.} Validation bit error over training for PixelSeal and VideoSeal under random augmentation (blue) and the learned adversary (orange). Across both backbones, CAT reaches lower validation bit error earlier, indicating that CAT identifies and improves upon informative training-time failures more effectively than random augmentation.}
    \label{fig:convergence_comparison}
\end{figure*}

\paragraph{CAT improves sample efficiency by focusing training on informative failures.}
Figure~\ref{fig:convergence_comparison} shows that CAT consistently reaches lower validation bit error earlier than random augmentation for both PixelSeal and VideoSeal, and that this advantage persists from the single-step setting to the compositional setting. The gap is especially meaningful because it appears throughout training rather than only at the end: the learned adversary exposes failure cases that are already relevant to the model's current weaknesses, whereas random augmentation spends many updates on transformations that are either too easy or insufficiently targeted. This makes optimization more sample efficient, not merely more aggressive. The result is particularly important in the compositional attack regime, where random augmentation must discover useful short attack compositions by chance, while CAT can adaptively concentrate on the sequences that actually break decoding. Because our method reuses the standard training pass instead of introducing a separate inner attack search, the earlier drop in validation bit error reflects a genuine efficiency gain in how robust supervision is allocated during training.

\subsection{Image Quality}

Table~\ref{tab:quality} shows that introducing the learned adversary does not materially degrade the visual fidelity of the watermarked images. Across VideoSeal 0.0, VideoSeal 1.0, and PixelSeal, CAT remains very close to the compute-matched random-augmentation baseline on PSNR, SSIM, MS-SSIM, and LPIPS, with only small trade-offs that are minor relative to the robustness gains reported above. In particular, perceptual quality remains strong for all CAT-trained models, LPIPS is unchanged or improved for both VideoSeal variants, and PixelSeal maintains essentially identical LPIPS under CAT. Taken together, these results suggest that CAT improves robustness primarily by reallocating training pressure toward difficult corruptions rather than by writing a more visible or intrusive watermark.

\begin{table}[ht]
\caption{\textbf{CAT preserves visual quality while improving robustness.} Image-quality results on SA-1B and DIV2K, reported using PSNR, SSIM, MS-SSIM, and LPIPS. Across VideoSeal 0.0, VideoSeal 1.0, and PixelSeal, the ``+ CAT'' models remain very close to their compute-matched random-augmentation baselines.}
\centering
\scriptsize
\resizebox{0.7\linewidth}{!}{%
\begin{tabular}{lcccccccc}
\toprule
 & \multicolumn{4}{c}{\textbf{SA-1B}} & \multicolumn{4}{c}{\textbf{DIV2K}} \\
\cmidrule(lr){2-5} \cmidrule(lr){6-9}
& PSNR$\uparrow$ & SSIM$\uparrow$ & MS-SSIM$\uparrow$ & LPIPS$\downarrow$ & PSNR$\uparrow$ & SSIM$\uparrow$ & MS-SSIM$\uparrow$ & LPIPS$\downarrow$ \\
\midrule
InvisMark & 48.77 & 0.9955 & 0.9964 & 0.0018 & 49.11 & 0.9943 & 0.9960 & 0.0016 \\
TrustMark & 41.37 & 0.9943 & 0.9917 & 0.0029 & 41.19 & 0.9935 & 0.9919 & 0.0027 \\
MBRS & 45.58 & 0.9959 & 0.9965 & 0.0032 & 45.22 & 0.9954 & 0.9966 & 0.0034 \\
\midrule
VideoSeal 0.0 & 42.50 & 0.9934 & 0.9949 & 0.0049 & 42.11 & 0.9910 & 0.9944 & 0.0057 \\
\quad + CAT & 42.21 & 0.9935 & 0.9953 & 0.0040 & 41.81 & 0.9911 & 0.9947 & 0.0046 \\
\cmidrule(lr){1-9}
VideoSeal 1.0 & 42.58 & 0.9936 & 0.9950 & 0.0046 & 42.19 & 0.9913 & 0.9945 & 0.0053 \\
\quad + CAT & 42.17 & 0.9934 & 0.9951 & 0.0039 & 41.75 & 0.9909 & 0.9946 & 0.0045 \\
\cmidrule(lr){1-9}
PixelSeal & 43.22 & 0.9958 & 0.9965 & 0.0021 & 42.71 & 0.9940 & 0.9961 & 0.0023 \\
\quad + CAT & 42.64 & 0.9956 & 0.9963 & 0.0021 & 42.17 & 0.9937 & 0.9958 & 0.0024 \\
\bottomrule
\end{tabular}%
}
\label{tab:quality}
\end{table}

\subsection{Ablations}

We ablate three design choices in CAT to isolate the source of the robustness gains. First, we remove the entropy regularization term from the adversary objective, which tests whether maintaining a diverse attack policy is necessary for learning broad robustness rather than overfitting to a narrow set of destructive sequences. Second, we replace the frozen DINOv3 feature backbone in the adversary with a ResNet18 backbone, testing whether the stronger visual representation is important for choosing informative sequential attacks. Third, we replace our straight-through Gumbel-Softmax attack selector with a multi-armed bandit adversary based on upper confidence bounds (UCB), which provides a non-differentiable exploration strategy for adversarial attack selection.

\paragraph{Entropy regularization and differentiable attack selection are necessary for robust watermarking.}
Tables~\ref{tab:ablation_d1} and~\ref{tab:ablation_d2} show that removing entropy regularization causes the largest collapse in robustness, consistent with the adversary prematurely concentrating on a narrow attack subset. Replacing DINOv3 with ResNet18 also weakens performance, indicating that stronger frozen features help the controller identify more informative failure cases. Finally, substituting Gumbel-Softmax with the UCB bandit adversary narrows the gap less severely than the other ablations, but still underperforms the full method, suggesting that end-to-end differentiable attack selection is more effective than a bandit-style exploration heuristic for this setting.

\begin{table}[htbp]
\centering
\caption{Ablation study on single-step attacks on SA-1B. Each cell reports bit accuracy and watermark capacity for one augmentation family.}
\label{tab:ablation_d1}
\resizebox{\textwidth}{!}{
\begin{tabular}{l*{6}{cc}}
\toprule
Method & \multicolumn{2}{c}{Identity} & \multicolumn{2}{c}{Value} & \multicolumn{2}{c}{Compression} & \multicolumn{2}{c}{Geometric} & \multicolumn{2}{c}{Combined} & \multicolumn{2}{c}{Overall} \\
\cmidrule(lr){2-3}\cmidrule(lr){4-5}\cmidrule(lr){6-7}\cmidrule(lr){8-9}\cmidrule(lr){10-11}\cmidrule(lr){12-13}
 & Bit acc. $(\uparrow)$ & Capacity $(\uparrow)$ & Bit acc. $(\uparrow)$ & Capacity $(\uparrow)$ & Bit acc. $(\uparrow)$ & Capacity $(\uparrow)$ & Bit acc. $(\uparrow)$ & Capacity $(\uparrow)$ & Bit acc. $(\uparrow)$ & Capacity $(\uparrow)$ & Bit acc. $(\uparrow)$ & Capacity $(\uparrow)$ \\
\midrule
\textbf{CAT} & 0.986 & 117.73 & 0.944 & 99.88 & 0.981 & 114.78 & 0.837 & 74.93 & 0.973 & 109.36 & \textbf{0.925} & \textbf{91.91} \\
\quad - Entropy Reg. & 0.987 & 116.32 & 0.940 & 95.50 & 0.985 & 115.24 & 0.678 & 36.62 & 0.977 & 110.03 & 0.766 & 54.73 \\
\quad + ResNet18 & 0.969 & 106.40 & 0.923 & 88.22 & 0.961 & 102.47 & 0.776 & 55.04 & 0.950 & 96.16 & 0.856 & 68.72 \\
\quad + UCB & 0.997 & 125.09 & 0.949 & 103.02 & 0.994 & 123.63 & 0.808 & 70.69 & 0.979 & 114.24 & 0.893 & 87.86 \\
\bottomrule
\end{tabular}
}
\end{table}
\begin{table}[htbp]
\caption{Ablation study on compositional attacks on SA-1B. Each cell reports bit accuracy and watermark capacity for a pair of augmentation families.}
\centering
\resizebox{\textwidth}{!}{
\begin{tabular}{l*{7}{cc}}
\toprule
Method & \multicolumn{2}{c}{Val+Val} & \multicolumn{2}{c}{Val+Comp} & \multicolumn{2}{c}{Val+Geom} & \multicolumn{2}{c}{Comp+Comp} & \multicolumn{2}{c}{Comp+Geom} & \multicolumn{2}{c}{Geom+Geom} & \multicolumn{2}{c}{Overall} \\
\cmidrule(lr){2-3}\cmidrule(lr){4-5}\cmidrule(lr){6-7}\cmidrule(lr){8-9}\cmidrule(lr){10-11}\cmidrule(lr){12-13}\cmidrule(lr){14-15}
 & Bit acc. $(\uparrow)$ & Capacity $(\uparrow)$ & Bit acc. $(\uparrow)$ & Capacity $(\uparrow)$ & Bit acc. $(\uparrow)$ & Capacity $(\uparrow)$ & Bit acc. $(\uparrow)$ & Capacity $(\uparrow)$ & Bit acc. $(\uparrow)$ & Capacity $(\uparrow)$ & Bit acc. $(\uparrow)$ & Capacity $(\uparrow)$ & Bit acc. $(\uparrow)$ & Capacity $(\uparrow)$ \\
\midrule
\textbf{CAT} & 0.974 & 113.81 & 0.990 & 121.27 & 0.964 & 107.65 & 0.998 & 126.29 & 0.982 & 116.19 & 0.928 & 100.21 & \textbf{0.965} & \textbf{107.73} \\
- Entropy Reg. & 0.504 & 0.71 & 0.505 & 0.71 & 0.503 & 0.72 & 0.505 & 0.70 & 0.503 & 0.72 & 0.503 & 0.73 & 0.503 & 0.72 \\
+ ResNet18 & 0.974 & 112.73 & 0.986 & 118.54 & 0.963 & 106.29 & 0.995 & 124.10 & 0.979 & 113.81 & 0.923 & 97.12 & 0.961 & 105.36 \\
+ UCB & 0.965 & 106.95 & 0.978 & 113.30 & 0.876 & 79.10 & 0.988 & 119.05 & 0.893 & 85.13 & 0.817 & 65.51 & 0.868 & 75.88 \\
\bottomrule
\end{tabular}
}
\label{tab:ablation_d2}
\end{table}
\section{Limitations and Conclusion}
While our study demonstrates the benefits of adaptive attack sequences for improving robustness and sample efficiency, one limitation is that we focus on short-horizon differentiable attacks within a bounded transformation space. As a result, our evaluation does not fully capture the broader range of real-world edits that watermarked content may encounter after deployment. In particular, longer editing chains and higher-level semantic manipulations remain outside the scope of our current study. Extending adaptive compositional adversaries to these settings is an important direction for future work. Overall, our results support the view that robust visual watermarking is fundamentally a robustness problem over structured and compositional transformations rather than isolated perturbations, and they suggest that future watermarking systems may benefit less from indiscriminately increasing augmentation diversity and more from learning to target the structured failure modes that dominate practical settings.

\section*{Acknowledgements}
 Satheesh, Panaitescu-Liess, Xu, Cai, and Huang F. are supported by DARPA HR001124S0029-AIQ-FP-019, National Science Foundation TRAILS Institute (2229885). Private support was provided by Open Philanthropy and Apple. The Authors acknowledge the National Artificial Intelligence Research Resource (NAIRR) Pilot for contributing to this research result.
\clearpage
\bibliography{main}

\appendix
\section{Extended Related Work}\label{sec:extended-related-work}

\paragraph{Traditional Watermarking.}
Early watermarking methods embedded signals in spatial or frequency domains such as DCT~\citep{piva1997dct} and DWT~\citep{barni2001improved, xia1998wavelet}, relying on hand-crafted rules designed for specific distortion types. While effective under controlled conditions, these approaches generally lack adaptability to diverse transformations. This motivated the development of model-based methods that incorporate differentiable transformations directly into training.

\paragraph{Model-based Image Watermarking.}
Deep neural networks have become the dominant approach to watermarking, largely following the encoder--noise--decoder framework introduced by HiDDeN~\citep{zhu2018hidden}. In this formulation, the noise layer simulates distortions during training to improve robustness. Subsequent work extends this framework with architectural and training improvements, including residual diffusion embedding~\citep{ahmadi2020redmark}, attention mechanisms~\citep{zhang2022digital}, and improved robustness to compression via mini-batch mixing of real and simulated JPEG~\citep{jia2021mbrs}. More recent approaches address resolution-independent embedding~\citep{bui2025trustmark, xu2025invismark}, tamper localization~\citep{zhang2024editguard}, and robustness to diffusion-based purification~\citep{lu2024robust, nie2022diffusion, zhao2024invisible, saberi2023robustness}. In parallel, in-generation watermarking methods such as Stable Signature~\citep{fernandez2023stable}, Tree-Ring~\citep{wen2023tree}, and WMAR~\citep{jovanovic2025watermarking} embed watermarks directly during image synthesis. Many post-processing methods, including PixelSeal~\citep{souvcek2025pixel}, TrustMark~\citep{bui2025trustmark}, InvisMark~\citep{xu2025invismark}, and MBRS~\citep{jia2021mbrs}, are trained using noise layers composed of fixed or randomly sampled augmentations.

\paragraph{Model-based Video Watermarking.}
Video watermarking extends these ideas to the temporal domain but introduces additional computational challenges. Early approaches such as VStegNet~\citep{mishra2019vstegnet} and RivaGAN~\citep{zhang2019robust} adapted image-based architectures to video, often operating on full spatio-temporal tensors. Later works, including DVMark~\citep{luo2023dvmark}, VHNet~\citep{shen2023vhnet}, RC-VWN~\citep{chen2024robust}, and StegaVideo~\citep{hu2024stegavideo}, introduced improved spatio-temporal architectures, multiscale designs, and differentiable approximations of compression to enhance robustness and efficiency. Alternative approaches such as ItoV~\citep{ye2023itov} and VideoSeal~\citep{fernandez2024video} reduce computational complexity by folding the temporal dimension into the channel dimension, enabling the use of image-based architectures for video watermarking. PixelSeal~\citep{souvcek2025pixel} also extends to video, using adversarial training with a discriminator to enforce perceptual quality.

\paragraph{Composite Adversaries.} 
Adversarial training against a single perturbation type, most commonly an $\ell_p$-norm ball, has been studied thoroughly~\citep{goodfellow2014explaining, madry2017towards, szegedy2013intriguing, carlini2017towards}. However, robustness to a particular threat model does not generalize to overall robustness against all attacks. Several works have shown that training against one threat model can actually increase vulnerability to others~\citep{tramer2019adversarial, maini2020adversarial}. This has motivated a line of work on adversarial training against a union or composition of multiple perturbation types~\citep{tramer2019adversarial, maini2020adversarial, croce2020reliable, laidlaw2020perceptual, rowe2022closer}. Most closely related to our setting, \citet{hsiung2023towards} proposed generalized adversarial training (GAT) over composite semantic perturbations, including combinations of hue, saturation, brightness, contrast, and rotation. In parallel, some literature has explored learning augmentation policies rather than fixing them by hand~\citep{cubuk2019autoaugment, li2020differentiable, liu2021direct, ho2019population, muller2021trivialaugment}. \citet{zhang2019adversarial} further refined augmentation policy search into a min-max game between a target network and an augmentation network, jointly optimizing both. Rather than training for classification robustness or generalization accuracy, our work involves training an adversary designed to break a watermark encoder-extractor pair, operating over a structured space of post-processing operations relevant to the image and video distribution shift problem. Furthermore, unlike GAT \citep{hsiung2023towards}, which schedules attack order as a discrete search problem, our adversary is a continuous, sequential, differentiable agent trained end-to-end via gradient flow, enabling it to discover attack compositions that are not enumerable by hand.

\section{Further Implementation Details}\label{sec:further-implementation-details}

\subsection{Model Architectures and Training Setup}\label{sec:model-training-details}

\paragraph{Watermarking backbones and adversary architecture.}
The embedder and extractor use the same architectures as in \citep{souvcek2025pixel}: the embedder is a U-Net with 8 residual blocks and 16 base channels with multipliers $[1, 2, 4, 8]$, operating on the YUV luminance channel with group normalization and GELU activations, while the extractor uses a ConvNeXt-Tiny backbone with depths $[3, 3, 9, 3]$ and channel dimensions $[96, 192, 384, 768]$, followed by a pixel decoder for message extraction. At each adversarial step $t$, visual features are extracted with a frozen DINOv3 ViT-S/16 backbone \citep{simeoni2025dinov3} and projected by a two-layer MLP before being passed to the GRU controller in Eq.~\ref{eq:gru}. A two-layer MLP attack head maps the hidden state to logits $\ell_t \in \mathbb{R}^K$ for Gumbel-Softmax selection. The DINOv3 backbone is frozen to avoid representation collapse, and the adversary only finetunes the GRU and MLP parameters for attack selection.

\paragraph{Training hyperparameters.}
For all methods, both the watermark model $(\theta, \psi)$ and adversary $\phi$ are trained on a 50{,}000-image subset of SA-1B \citep{kirillov2023segment} for 500 epochs with an effective batch size of 50 (batch size 25 with gradient accumulation over 2 steps) using 4 NVIDIA A4000 GPUs, yielding approximately 120K gradient steps. We use AdamW with learning rate $5\times10^{-4}$, a cosine schedule decaying to $10^{-6}$, and 5 epochs of linear warm-up before activating the adversary. The watermark scaling factor follows a cosine decay from $\alpha_0 = 1.0$ to $\alpha_1 = 0.2$ over the final 200 epochs. Loss weights are $\lambda_{\text{dec}} = 1.0$ for message decoding and $\lambda_i = 0.1$ for imperceptibility, both applied in YUV space. Unless noted otherwise, the adversary uses depth $T \in \{1, 2\}$, entropy coefficient $\lambda_{\text{ent}} = 0.1$, Gumbel temperature $\tau = 1.0$, and hidden dimension $d_h = 384$.

\subsection{Entropy Regularization and Optimization Details}\label{sec:joint_optimization}

\paragraph{Entropy decomposition and single-pass optimization.}
Since the policy factorizes as $\pi_\phi(a) = \prod_t \pi_{\phi,t}(a_t)$, the joint entropy decomposes as $\mathcal{H}(\pi_\phi) = \sum_t \mathcal{H}(\pi_{\phi,t})$. Defining the per-step softmax distribution $p_t^{(i)} = \operatorname{softmax}(\ell_t / \tau_{\text{ent}})_i$ and entropy $H_t = -\sum_{i=1}^{K} p_t^{(i)} \log p_t^{(i)}$, the adversary objective becomes
\begin{align}
    \mathcal{J}_{\mathrm{adv}}(\phi)
    &= \mathbb{E}_{a \sim \pi_\phi}\bigl[\mathcal{L}_{\mathrm{msg}}\bigl(D_\psi(x_T),\, m\bigr)\bigr]
    + \lambda_{\text{ent}} \sum_{t=0}^{T-1} H_t, \\
    H_t
    &= -\sum_{i=1}^{K} p_t^{(i)} \log p_t^{(i)}.
\end{align}
This formulation makes the entropy bonus straightforward to compute from the controller logits at each attack step. In practice, it keeps the controller exploratory early in training, avoids premature collapse to a trivially destructive strategy, and improves robustness beyond a single dominant failure mode. We then optimize the watermark model and adversary jointly in a single pass rather than through a separate inner maximization loop. Concretely, we implement the entropy bonus as an entropy loss $\mathcal{L}_{\text{ent}} := -\sum_{t=0}^{T-1} H_t$, so the adversary receives the sign-reversed task gradient
\begin{equation}
    \nabla_\phi \mathcal{L}_{\text{adv}}
    = -\nabla_\phi \mathcal{L}_{\text{msg}}
    + \lambda_{\text{ent}}\, \nabla_\phi \mathcal{L}_{\text{ent}}.
    \label{eq:grad_negation}
\end{equation}
This lets the watermark model and the adversary share a single computational graph per iteration instead of alternating between separate inner and outer optimization loops.

\subsection{Evaluation Methodology}\label{sec:evaluation-methodology}

\paragraph{Metrics.}
We evaluate robustness primarily with \emph{bit accuracy} and \emph{capacity}. Bit accuracy is the fraction of watermark bits recovered correctly after augmentation, computed by thresholding the extractor outputs at $0.5$ and averaging the resulting bitwise matches against the ground-truth message. Capacity is reported through the corresponding binary symmetric channel estimate: for a bit recovered with accuracy $p$, the per-bit capacity is $1 - H(p)$, where $H(p) = -p\log_2 p - (1-p)\log_2(1-p)$ is the Bernoulli entropy, and the total capacity multiplies this quantity by the payload size. We additionally compute a one-sided binomial-test $p$-value against the null hypothesis of random guessing ($p=0.5$) to verify that decoding performance is statistically above chance. For perceptual distortion, image experiments report PSNR, SSIM, MS-SSIM, and LPIPS, while video experiments report VMAF and BD-rate.

\paragraph{Evaluation pipeline.}
Each sample is first watermarked by the model, producing a marked image or video together with the target message. We then apply either a single-step attack, a composed augmentation, or a compositional attack pair, depending on the experiment. The extractor is run on the transformed content, and the recovered message is compared against the original payload to compute bit accuracy, capacity, and significance statistics. For videos, framewise predictions are aggregated by averaging before thresholding, which improves stability when the watermark signal is distributed across time.

\paragraph{Augmentation protocol.}
For images, the evaluation suite covers geometric, photometric, and compression perturbations including rotation, resize, crop, perspective distortion, horizontal flip, brightness, contrast, hue, grayscale conversion, Gaussian blur, and JPEG compression. In the compositional setting, we additionally evaluate all ordered pairs of the base image augmentations at reduced strength to measure robustness under weak but composed distortions. For videos, we extend the same families with codec-based and temporal corruptions, including H.264/H.265 compression, playback-speed changes, frame dropping, temporal averaging, and temporal reordering. We also report results for stronger composed chains such as compression $\rightarrow$ crop $\rightarrow$ brightness to test robustness under realistic sequential edits.

\begin{table*}[t]
    \centering
    \small
    \setlength{\tabcolsep}{4pt}
    \caption{Evaluation-time augmentation grids for image and video experiments. Single-step attacks use the full parameter grid for each augmentation. Compositional attacks enumerate unordered pairs of augmentations and use the reduced parameter grids listed here; for binary augmentations, the same toggle is used in both settings.}
    \label{tab:augmentation_grids}
    \resizebox{\textwidth}{!}{
    \begin{tabular}{l l p{6.2cm} p{6.4cm}}
        \toprule
        Domain & Augmentation & Single-step parameters & Compositional parameters \\
        \midrule
        Image & Rotate & $\{5^\circ, 10^\circ, 30^\circ, 45^\circ, 90^\circ\}$ & $\{5^\circ, 10^\circ\}$ \\
        Image & Resize & \shortstack[l]{$\{0.32, 0.45, 0.55, 0.63, 0.71,$\\$0.77, 0.84, 0.89, 0.95, 1.00\}$} & $\{0.71, 0.77, 0.84, 0.89\}$ \\
        Image & Crop & \shortstack[l]{$\{0.32, 0.45, 0.55, 0.63, 0.71,$\\$0.77, 0.84, 0.89, 0.95, 1.00\}$} & $\{ 0.71, 0.77, 0.84, 0.89\}$ \\
        Image & Perspective & $\{0.1, 0.2, 0.3, 0.4, 0.5, 0.6, 0.7, 0.8\}$ & $\{0.1, 0.2, 0.3, 0.4\}$ \\
        Image & Horizontal flip & $\{\text{off}, \text{on}\}$ & --- \\
        Image & Brightness & $\{0.1, 0.25, 0.5, 0.75, 1.0, 1.25, 1.5, 1.75, 2.0\}$ & $\{0.5, 0.75, 1.25, 1.5\}$ \\
        Image & Contrast & $\{0.1, 0.25, 0.5, 0.75, 1.0, 1.25, 1.5, 1.75, 2.0\}$ & $\{0.5, 0.75, 1.25, 1.5\}$ \\
        Image & Hue & \shortstack[l]{$\{-0.4, -0.3, -0.2, -0.1, 0.0,$\\$0.1, 0.2, 0.3, 0.4, 0.5\}$} & $\{-0.3, -0.2, -0.1, 0.1, 0.2, 0.3\}$ \\
        Image & Grayscale & $\{\text{off}, \text{on}\}$ & --- \\
        Image & Gaussian blur & $\{3, 5, 9, 13, 17\}$ px kernel & $\{3, 5, 9\}$ px kernel \\
        Image & JPEG & $\{40, 50, 60, 70, 80, 90\}$ quality & $\{60, 70, 80\}$ quality \\
        \midrule
        Video & Rotate & $\{5^\circ, 10^\circ, 30^\circ, 45^\circ, 90^\circ\}$ & $\{5^\circ, 10^\circ\}$ \\
        Video & Resize & \shortstack[l]{$\{0.32, 0.45, 0.55, 0.63, 0.71,$\\$0.77, 0.84, 0.89, 0.95, 1.00\}$} & $\{0.71, 0.77, 0.84, 0.89\}$ \\
        Video & Crop & \shortstack[l]{$\{0.32, 0.45, 0.55, 0.63, 0.71,$\\$0.77, 0.84, 0.89, 0.95, 1.00\}$} & $\{ 0.71, 0.77, 0.84, 0.89\}$ \\
        Video & Perspective & $\{0.1, 0.2, 0.3, 0.4, 0.5, 0.6, 0.7, 0.8\}$ & $\{0.1, 0.2, 0.3, 0.4, 0.5, 0.6\}$ \\
        Video & Horizontal flip & $\{\text{off}, \text{on}\}$ & --- \\
        Video & Brightness & $\{0.1, 0.25, 0.5, 0.75, 1.0, 1.25, 1.5, 1.75, 2.0\}$ & $\{0.5, 0.75, 1.25, 1.5\}$ \\
        Video & Contrast & $\{0.1, 0.25, 0.5, 0.75, 1.0, 1.25, 1.5, 1.75, 2.0\}$ & $\{0.5, 0.75, 1.25, 1.5\}$ \\
        Video & Hue & \shortstack[l]{$\{-0.4, -0.3, -0.2, -0.1, 0.0,$\\$0.1, 0.2, 0.3, 0.4, 0.5\}$} & $\{-0.2, -0.1, 0.1, 0.2\}$ \\
        Video & Saturation & $\{0.5, 1.5\}$ & $\{0.5, 0.75, 1.25, 1.5\}$ \\
        Video & Grayscale & $\{\text{off}, \text{on}\}$ & --- \\
        Video & Gaussian blur & $\{3, 5, 9, 13, 17\}$ px kernel & $\{3, 5, 9\}$ px kernel \\
        Video & JPEG & $\{40, 50, 60, 70, 80, 90\}$ quality & $\{60, 70, 80\}$ quality \\
        Video & H.264 & $\{9, 16, 23, 30, 40, 50\}$ CRF & $\{28, 34\}$ CRF \\
        Video & H.265 & $\{9, 16, 23, 30, 40, 50\}$ CRF & --- \\
        Video & Speed change & $\{0.5\times, 0.75\times, 1.25\times, 2.0\times\}$ & $\{0.75\times, 1.25\times\}$ \\
        Video & Drop frame & $\{0.0, 0.15, 0.3\}$ drop prob. & $\{0.0, 0.15\}$ drop prob. \\
        Video & Window averaging & $\{0.2, 0.4, 0.6, 0.8\}$ & $\{0.2, 0.4\}$ \\
        Video & Temporal reorder & $\{0.0, 0.25, 0.5, 0.75, 1.0\}$ & $\{0.25, 0.5\}$ \\
        \bottomrule
    \end{tabular}}
\end{table*}

\paragraph{Datasets and implementation details.}
Image evaluation uses SA-1B for in-distribution testing together with multiple out-of-distribution image sets (CLIC, DIV2K, and MetFace), while video evaluation is reported on Movie-Gen-Bench \citep{polyak2024movie} for in-distribution testing and SA-V \citep{ravi2024sam} for out-of-distribution testing. Unless otherwise specified, all listed augmentations are evaluated independently, and the compositional appendix results additionally include the full pairwise augmentation set. LPIPS uses the AlexNet backbone, and video quality metrics are computed with the same settings across methods to ensure a fair comparison.
\section{Additional OOD Image Results}\label{sec:ood-image-results}

\paragraph{CAT's robustness gains persist under distribution shift.}
Tables~\ref{tab:results_depth1_ood_additional} and~\ref{tab:results_depth2_ood_additional} extend the main results to additional OOD datasets, namely DIV2K \citep{agustsson2017ntire} and MetFace. The same overall pattern remains visible: CAT improves overall robustness relative to the compute-matched random-augmentation baseline across both datasets, with the clearest gains again appearing in watermark capacity. In the single-step setting, overall capacity rises from 102.28 to 119.08 on DIV2K and from 118.26 to 151.78 on MetFace for VideoSeal 1.0, while PixelSeal improves from 54.28 to 86.50 on DIV2K and from 65.54 to 104.96 on MetFace. These gains are much larger than the corresponding changes in average bit accuracy, indicating that CAT is making decoding reliably correct across a broader set of corruptions rather than producing isolated improvements on a few easy cases.

\paragraph{The gains are strongest on the hardest OOD corruption settings.}
The hardest OOD settings again show the largest benefits. In the single-step results, PixelSeal gains 59.4\% capacity on DIV2K and 60.1\% on MetFace, while VideoSeal 1.0 gains 16.4\% and 28.3\%, respectively. In the compositional OOD results, the same pattern holds on the harder mixed corruption pairs: on DIV2K, VideoSeal 1.0 improves from 130.74 to 165.63 on Comp+Comp and from 103.87 to 132.49 on Comp+Geom, while PixelSeal improves from 116.91 to 125.54 and from 105.55 to 114.86 on the same two settings. On MetFace, VideoSeal 1.0 improves from 138.93 to 184.98 on Comp+Comp and from 122.22 to 169.31 on Comp+Geom. This behavior is consistent with the main paper's hypothesis that adaptive compositional adversaries are most valuable when robustness depends on rare, structured corruptions rather than on isolated perturbations.

\paragraph{OOD generalization remains consistent with the main results.}
The appendix results show that CAT does not merely overfit to the in-distribution evaluation setting or to a narrow attack suite. Even when both the image distribution and the corruption patterns shift, the learned adversary continues to improve overall robustness for all three learned backbones in both the single-step and compositional settings. While the absolute difficulty varies across datasets and models, the largest gains again appear for the more vulnerable backbones, especially VideoSeal 1.0 and PixelSeal, reinforcing the conclusion that targeted compositional training improves robustness beyond the training distribution.


\begin{table}[htbp]
\caption{\textbf{The robustness gains from a single learned attack step persist across more challenging out-of-distribution datasets, with the largest improvements again appearing for VideoSeal 1.0 and PixelSeal.} Additional single-step attack results on OOD DIV2K and MetFace images. Each cell reports bit accuracy and capacity for identity, value, compression, geometric, combined, and overall evaluation settings. For VideoSeal and PixelSeal, the ``+ CAT'' rows use Compositional Adversarial Training, while the corresponding base rows use random augmentation training; higher is better for both metrics.}\label{tab:results_depth1_ood_additional}
\centering
\resizebox{\textwidth}{!}{
\begin{tabular}{l l*{6}{cc}}
\toprule
 &  & \multicolumn{2}{c}{Identity} & \multicolumn{2}{c}{Value} & \multicolumn{2}{c}{Compression} & \multicolumn{2}{c}{Geometric} & \multicolumn{2}{c}{Combined} & \multicolumn{2}{c}{Overall} \\
\cmidrule(lr){3-4}\cmidrule(lr){5-6}\cmidrule(lr){7-8}\cmidrule(lr){9-10}\cmidrule(lr){11-12}\cmidrule(lr){13-14}
 & Model (bits) & Bit acc. $(\uparrow)$ & Capacity $(\uparrow)$ & Bit acc. $(\uparrow)$ & Capacity $(\uparrow)$ & Bit acc. $(\uparrow)$ & Capacity $(\uparrow)$ & Bit acc. $(\uparrow)$ & Capacity $(\uparrow)$ & Bit acc. $(\uparrow)$ & Capacity $(\uparrow)$ & Bit acc. $(\uparrow)$ & Capacity $(\uparrow)$ \\
\midrule
\multirow{9}{*}{\rotatebox{90}{\textbf{DIV2k}}} & InvisMark (100) & 0.990 & 96.03 & 0.878 & 73.49 & 0.953 & 88.43 & 0.828 & 63.40 & 0.834 & 63.64 & 0.863 & 70.23 \\
 & TrustMark (100) & 0.991 & 97.80 & 0.951 & 85.26 & 0.893 & 77.98 & 0.751 & 49.54 & 0.987 & 96.54 & 0.885 & 75.06 \\
 & MBRS (256) & 0.982 & 238.04 & 0.911 & 181.85 & 0.878 & 184.94 & 0.651 & 77.28 & 0.944 & 204.49 & 0.828 & 153.33 \\
\cmidrule(lr){2-14}
\morecmidrules\cmidrule(lr){2-14}
 & VideoSeal 0.0 (96) & 0.993 & 92.24 & 0.976 & 86.31 & 0.972 & 83.96 & 0.934 & 72.17 & 0.987 & 89.56 & 0.952 & 77.98 \\
 & \quad + CAT & 0.996 & 93.57 & 0.974 & 85.63 & 0.983 & 87.71 & 0.951 & 77.59 & 0.990 & 90.70 & \textbf{0.963}$_{\textcolor{ForestGreen}{\footnotesize \uparrow 1.2\%}}$ & \textbf{81.67}$_{\textcolor{ForestGreen}{\footnotesize \uparrow 4.7\%}}$ \\
\cmidrule(lr){2-14}
 & VideoSeal 1.0 (256) & 0.893 & 131.86 & 0.873 & 120.14 & 0.864 & 114.63 & 0.813 & 90.08 & 0.884 & 125.72 & 0.838 & 102.28 \\
 & \quad + CAT & 0.934 & 169.64 & 0.848 & 125.18 & 0.883 & 138.64 & 0.821 & 108.05 & 0.905 & 150.51 & \textbf{0.842}$_{\textcolor{ForestGreen}{\footnotesize \uparrow 0.5\%}}$ & \textbf{119.08}$_{\textcolor{ForestGreen}{\footnotesize \uparrow 16.4\%}}$ \\
\cmidrule(lr){2-14}
 & PixelSeal (128) & 0.914 & 74.84 & 0.890 & 67.21 & 0.873 & 62.00 & 0.809 & 46.05 & 0.897 & 68.60 & 0.841 & 54.28 \\
 & \quad + CAT & 0.978 & 113.19 & 0.941 & 97.38 & 0.944 & 97.17 & 0.886 & 77.78 & 0.961 & 104.03 & \textbf{0.911}$_{\textcolor{ForestGreen}{\footnotesize \uparrow 8.3\%}}$ & \textbf{86.50}$_{\textcolor{ForestGreen}{\footnotesize \uparrow 59.4\%}}$ \\
\midrule
\multirow{9}{*}{\rotatebox{90}{\textbf{MetFace}}} & InvisMark (100) & 0.990 & 96.04 & 0.887 & 76.05 & 0.932 & 82.01 & 0.831 & 64.16 & 0.671 & 37.40 & 0.825 & 64.06 \\
 & TrustMark (100) & 1.000 & 99.98 & 0.979 & 93.44 & 0.904 & 80.36 & 0.762 & 51.84 & 0.996 & 98.36 & 0.900 & 78.87 \\
 & MBRS (256) & 0.995 & 250.90 & 0.951 & 216.87 & 0.891 & 196.88 & 0.656 & 81.63 & 0.930 & 199.79 & 0.839 & 165.16 \\
\cmidrule(lr){2-14}
\morecmidrules\cmidrule(lr){2-14}
 & VideoSeal 0.0 (96) & 0.999 & 95.54 & 0.996 & 93.85 & 0.989 & 90.87 & 0.966 & 83.45 & 0.972 & 86.33 & 0.976 & 87.06 \\
 & \quad + CAT & 1.000 & 95.87 & 0.993 & 92.94 & 0.995 & 93.54 & 0.974 & 86.89 & 0.969 & 86.01 & \textbf{0.981}$_{\textcolor{ForestGreen}{\footnotesize \uparrow 0.5\%}}$ & \textbf{89.31}$_{\textcolor{ForestGreen}{\footnotesize \uparrow 2.6\%}}$ \\
\cmidrule(lr){2-14}
 & VideoSeal 1.0 (256) & 0.904 & 139.36 & 0.897 & 134.84 & 0.886 & 128.26 & 0.848 & 108.87 & 0.869 & 121.09 & 0.866 & 118.26 \\
 & \quad + CAT & 0.954 & 187.91 & 0.902 & 158.54 & 0.925 & 168.65 & 0.882 & 144.17 & 0.875 & 147.32 & \textbf{0.894}$_{\textcolor{ForestGreen}{\footnotesize \uparrow 3.2\%}}$ & \textbf{151.78}$_{\textcolor{ForestGreen}{\footnotesize \uparrow 28.3\%}}$ \\
\cmidrule(lr){2-14}
 & PixelSeal (128) & 0.928 & 80.54 & 0.919 & 77.52 & 0.902 & 71.69 & 0.858 & 59.19 & 0.877 & 66.62 & 0.880 & 65.54 \\
 & \quad + CAT & 0.996 & 124.24 & 0.980 & 116.02 & 0.978 & 114.12 & 0.936 & 98.04 & 0.954 & 104.60 & \textbf{0.954}$_{\textcolor{ForestGreen}{\footnotesize \uparrow 8.4\%}}$ & \textbf{104.96}$_{\textcolor{ForestGreen}{\footnotesize \uparrow 60.1\%}}$ \\
\bottomrule
\end{tabular}
}
\end{table}

\begin{table}[htbp]
\caption{\textbf{Under compositional attacks, CAT yields its clearest OOD gains, with especially large improvements for VideoSeal 1.0 and PixelSeal on hard mixed and repeated attack pairs.} Additional out-of-distribution image results on DIV2K and MetFace, where the adversary applies a compositional two-step attack sequence. Each cell reports bit accuracy and capacity for pairwise compositions of value, compression, and geometric augmentation families, plus the overall average; higher is better for both metrics. For VideoSeal 0.0/1.0 and PixelSeal, the ``+ CAT'' rows use Compositional Adversarial Training, while the corresponding base rows use random augmentation training.}\label{tab:results_depth2_ood_additional}
\centering
\resizebox{\textwidth}{!}{
\begin{tabular}{l l*{7}{cc}}
\toprule
 &  & \multicolumn{2}{c}{Val+Val} & \multicolumn{2}{c}{Val+Comp} & \multicolumn{2}{c}{Val+Geom} & \multicolumn{2}{c}{Comp+Comp} & \multicolumn{2}{c}{Comp+Geom} & \multicolumn{2}{c}{Geom+Geom} & \multicolumn{2}{c}{Overall} \\
\cmidrule(lr){3-4}\cmidrule(lr){5-6}\cmidrule(lr){7-8}\cmidrule(lr){9-10}\cmidrule(lr){11-12}\cmidrule(lr){13-14}\cmidrule(lr){15-16}
 & Model (bits) & Bit acc. $(\uparrow)$ & Capacity $(\uparrow)$ & Bit acc. $(\uparrow)$ & Capacity $(\uparrow)$ & Bit acc. $(\uparrow)$ & Capacity $(\uparrow)$ & Bit acc. $(\uparrow)$ & Capacity $(\uparrow)$ & Bit acc. $(\uparrow)$ & Capacity $(\uparrow)$ & Bit acc. $(\uparrow)$ & Capacity $(\uparrow)$ & Bit acc. $(\uparrow)$ & Capacity $(\uparrow)$ \\
\midrule
\multirow{9}{*}{\rotatebox{90}{\textbf{DIV2k}}} & InvisMark (100) & 0.905 & 73.01 & 0.652 & 27.04 & 0.878 & 69.35 & 0.470 & 0.88 & 0.600 & 21.51 & 0.889 & 70.78 & 0.814 & 58.39 \\
 & TrustMark (100) & 0.954 & 85.92 & 0.961 & 87.50 & 0.782 & 52.14 & 0.965 & 88.47 & 0.771 & 49.21 & 0.703 & 35.58 & 0.830 & 61.32 \\
 & MBRS (256) & 0.914 & 179.76 & 0.825 & 123.78 & 0.598 & 38.69 & 0.752 & 79.13 & 0.550 & 13.86 & 0.496 & 0.97 & 0.649 & 58.04 \\
\cmidrule(lr){2-16}
\morecmidrules\cmidrule(lr){2-16}
 & VideoSeal 0.0 (96) & 0.974 & 84.53 & 0.985 & 88.28 & 0.962 & 78.64 & 0.991 & 90.92 & 0.972 & 82.37 & 0.928 & 72.38 & 0.961 & 78.19 \\
 & \quad + CAT & 0.986 & 89.32 & 0.993 & 92.33 & 0.974 & 84.32 & 0.997 & 94.43 & 0.985 & 88.45 & 0.928 & 75.53 & \textbf{0.973}$_{\textcolor{ForestGreen}{\footnotesize \uparrow 1.2\%}}$ & \textbf{83.78}$_{\textcolor{ForestGreen}{\footnotesize \uparrow 7.1\%}}$ \\
\cmidrule(lr){2-16}
 & VideoSeal 1.0 (256) & 0.869 & 118.04 & 0.881 & 124.43 & 0.819 & 91.54 & 0.891 & 130.74 & 0.848 & 103.87 & 0.789 & 82.20 & \textbf{0.820} & \textbf{91.83} \\
 & \quad + CAT & 0.832 & 117.82 & 0.882 & 140.06 & 0.814 & 104.11 & 0.928 & 165.63 & 0.877 & 132.49 & 0.810 & 104.21 & \textbf{0.813}$_{\textcolor{ForestGreen}{\footnotesize \downarrow 1.0\%}}$ & \textbf{101.81}$_{\textcolor{ForestGreen}{\footnotesize \uparrow 10.9\%}}$ \\
\cmidrule(lr){2-16}
 & PixelSeal (128) & 0.966 & 107.50 & 0.977 & 112.44 & 0.955 & 100.71 & 0.985 & 116.91 & 0.966 & 105.55 & 0.919 & 92.30 & 0.952 & 99.06 \\
 & \quad + CAT & 0.975 & 114.22 & 0.990 & 121.24 & 0.965 & 107.56 & 0.997 & 125.54 & 0.980 & 114.86 & 0.926 & 99.15 & \textbf{0.965}$_{\textcolor{ForestGreen}{\footnotesize \uparrow 1.3\%}}$ & \textbf{107.41}$_{\textcolor{ForestGreen}{\footnotesize \uparrow 8.4\%}}$ \\
\midrule
\multirow{9}{*}{\rotatebox{90}{\textbf{MetFace}}} & InvisMark (100) & 0.948 & 84.25 & 0.675 & 31.31 & 0.905 & 75.43 & 0.482 & 0.96 & 0.610 & 24.17 & 0.899 & 73.69 & 0.837 & 63.48 \\
 & TrustMark (100) & 0.989 & 96.23 & 0.989 & 96.24 & 0.809 & 58.60 & 0.994 & 97.66 & 0.798 & 55.76 & 0.728 & 40.72 & 0.858 & 68.53 \\
 & MBRS (256) & 0.964 & 223.33 & 0.902 & 177.38 & 0.625 & 58.61 & 0.853 & 140.42 & 0.576 & 27.29 & 0.496 & 0.96 & 0.685 & 83.17 \\
\cmidrule(lr){2-16}
\morecmidrules\cmidrule(lr){2-16}
 & VideoSeal 0.0 (96) & 0.990 & 92.59 & 0.998 & 94.88 & 0.990 & 90.81 & 0.999 & 95.33 & 0.993 & 91.85 & 0.958 & 83.70 & 0.990 & 90.66 \\
 & \quad + CAT & 0.997 & 94.59 & 0.999 & 95.48 & 0.994 & 93.17 & 1.000 & 95.88 & 0.997 & 94.52 & 0.954 & 84.28 & \textbf{0.994}$_{\textcolor{ForestGreen}{\footnotesize \uparrow 0.4\%}}$ & \textbf{92.86}$_{\textcolor{ForestGreen}{\footnotesize \uparrow 2.4\%}}$ \\
\cmidrule(lr){2-16}
 & VideoSeal 1.0 (256) & 0.895 & 133.85 & 0.900 & 136.70 & 0.856 & 111.78 & 0.903 & 138.93 & 0.879 & 122.22 & 0.822 & 98.89 & 0.856 & 111.52 \\
 & \quad + CAT & 0.892 & 153.16 & 0.924 & 169.78 & 0.881 & 143.41 & 0.950 & 184.98 & 0.932 & 169.31 & 0.863 & 135.41 & \textbf{0.882}$_{\textcolor{ForestGreen}{\footnotesize \uparrow 3.0\%}}$ & \textbf{142.40}$_{\textcolor{ForestGreen}{\footnotesize \uparrow 27.7\%}}$ \\
\cmidrule(lr){2-16}
 & PixelSeal (128) & 0.989 & 122.16 & 0.996 & 125.05 & 0.988 & 119.42 & 0.998 & 125.92 & 0.990 & 120.62 & 0.955 & 109.48 & 0.987 & 118.72 \\
 & \quad + CAT & 0.986 & 121.38 & 0.998 & 126.42 & 0.987 & 119.81 & 1.000 & 127.77 & 0.994 & 123.34 & 0.946 & 108.64 & \textbf{0.987}$_{\textcolor{ForestGreen}{\footnotesize \uparrow 0.0\%}}$ & \textbf{119.49}$_{\textcolor{ForestGreen}{\footnotesize \uparrow 0.7\%}}$ \\
\bottomrule
\end{tabular}
}
\end{table}

\section{Payload Scaling}\label{sec:appendix-payload-scaling}

\begin{figure*}[t]
    \centering
    \includegraphics[width=0.96\textwidth]{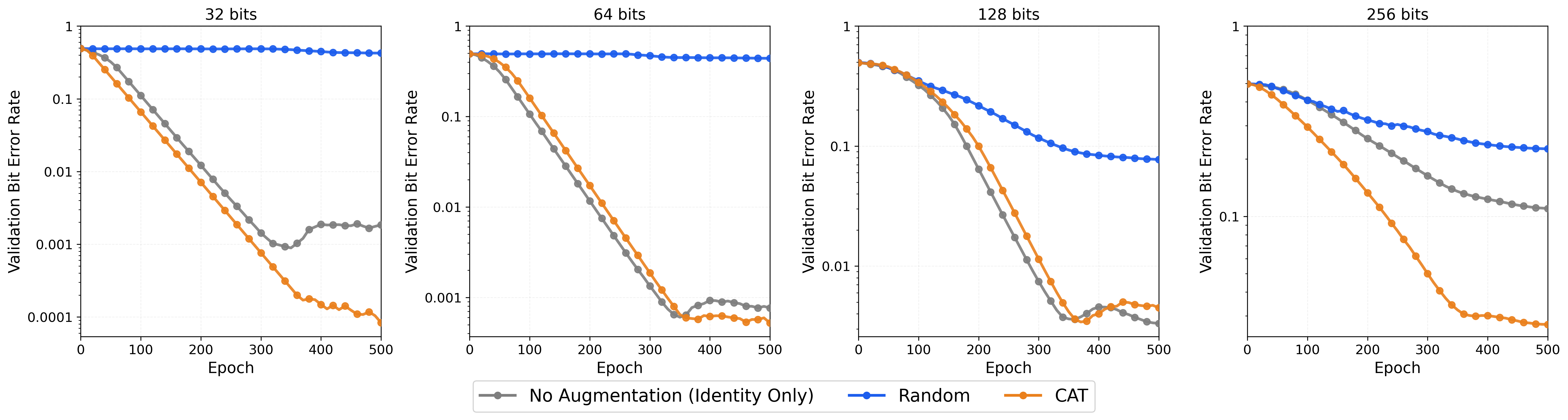}
    \caption{CAT improves PixelSeal training across payload sizes. Validation bit error rate over training is shown for 32-, 64-, 128-, and 256-bit payloads under no augmentation (gray), random augmentation (blue), and CAT (orange). CAT consistently drives the error lower than random augmentation and remains effective as payload increases, whereas random augmentation plateaus at substantially higher error.}
    \label{fig:payload_validation_convergence}
\end{figure*}

\paragraph{CAT improves convergence across payload sizes, with especially large gains at low payloads.}
Payload determines how many bits the model must embed and recover, and thus induces a trade-off between optimization difficulty and attainable robustness. Higher payloads are more difficult to train because the decoder must recover more bits under the same distortion budget, but they also correspond to higher effective capacity and therefore potentially stronger robustness once learned. By contrast, lower payloads are easier to fit under identity-only training, yet this regime is often more brittle because good clean-input performance can be achieved without learning invariance to realistic corruptions. An effective watermarking method should therefore provide both high capacity and robust performance across payload sizes, rather than relying on the deceptively easy low-payload identity regime. Figure~\ref{fig:payload_validation_convergence} makes this distinction clear. CAT improves convergence across the full range from 32 to 256 bits, with especially large gains at lower payloads, where it reduces validation error much more rapidly than random augmentation and in some cases even converges faster than the no-augmentation baseline. At the same time, CAT retains the lowest error on the augmented validation setting, showing that this speedup does not come from overfitting to the identity case.

Random augmentation performs particularly poorly at low payloads because the underlying clean task is already easy: the large gap between easy and hard attacks causes uniformly sampled augmentations to waste many updates on uninformative samples, yielding a brittle model that remains undertrained on the rarer corruptions that actually determine robustness. This effect is less visible in the identity-only baseline because the training objective is then aligned with the easiest validation case, but that apparent simplicity does not translate into robustness under realistic post-processing. CAT avoids this failure mode by adaptively concentrating training pressure on the transformations that remain challenging for the current model.

\section{Additional Video Results}\label{sec:video-results}

\begin{table}[htbp]
\caption{\textbf{Video watermarking single-step attack results on in-distribution Movie-Gen-Bench (ID) and out-of-distribution SAV-Test (OOD).} Each cell reports bit accuracy and capacity for identity, value, compression, geometric, video-temporal, and combined augmentation families. For VideoSeal and PixelSeal, the `+ CAT` rows use Compositional Adversarial Training while the base rows use random augmentation training; higher is better for both metrics.}
\centering
\resizebox{0.96\textwidth}{!}{
\begin{tabular}{l l*{7}{cc}}
\toprule
 &  & \multicolumn{2}{c}{Identity} & \multicolumn{2}{c}{Value} & \multicolumn{2}{c}{Compression} & \multicolumn{2}{c}{Geometric} & \multicolumn{2}{c}{Temporal} & \multicolumn{2}{c}{Combined} & \multicolumn{2}{c}{Overall} \\
\cmidrule(lr){3-4}\cmidrule(lr){5-6}\cmidrule(lr){7-8}\cmidrule(lr){9-10}\cmidrule(lr){11-12}\cmidrule(lr){13-14}\cmidrule(lr){15-16}
 & Model (bits) & Bit acc. $(\uparrow)$ & Capacity $(\uparrow)$ & Bit acc. $(\uparrow)$ & Capacity $(\uparrow)$ & Bit acc. $(\uparrow)$ & Capacity $(\uparrow)$ & Bit acc. $(\uparrow)$ & Capacity $(\uparrow)$ & Bit acc. $(\uparrow)$ & Capacity $(\uparrow)$ & Bit acc. $(\uparrow)$ & Capacity $(\uparrow)$ & Bit acc. $(\uparrow)$ & Capacity $(\uparrow)$ \\
\midrule
\multirow{7}{*}{\shortstack{\textbf{Movie-Gen-}\\\textbf{Bench (ID)}}} & RivaGAN (32) & 0.889 & 20.36 & 0.691 & 9.42 & 0.726 & 9.86 & 0.747 & 10.39 & 0.809 & 14.43 & 0.511 & 1.27 & 0.754 & 11.39 \\
\cmidrule(lr){2-16}
\morecmidrules\cmidrule(lr){2-16}
 & VideoSeal 0.0 (96) & 0.951 & 69.97 & 0.930 & 62.37 & 0.843 & 45.80 & 0.942 & 67.88 & 0.905 & 59.74 & 0.724 & 16.74 & 0.912 & 60.95 \\
 & \quad + CAT & 0.952 & 70.04 & 0.934 & 63.67 & 0.915 & 59.00 & 0.946 & 68.95 & 0.941 & 67.57 & 0.870 & 44.80 & \textbf{0.936}$_{\textcolor{ForestGreen}{\footnotesize \uparrow 2.6\%}}$ & \textbf{65.76}$_{\textcolor{ForestGreen}{\footnotesize \uparrow 7.9\%}}$ \\
\cmidrule(lr){2-16}
 & VideoSeal 1.0 (256) & 0.861 & 107.64 & 0.823 & 87.28 & 0.742 & 65.12 & 0.848 & 103.34 & 0.809 & 90.11 & 0.601 & 9.66 & 0.815 & 90.48 \\
 & \quad + CAT & 0.849 & 100.44 & 0.829 & 89.90 & 0.828 & 89.43 & 0.843 & 98.74 & 0.847 & 100.01 & 0.792 & 69.22 & \textbf{0.838}$_{\textcolor{ForestGreen}{\footnotesize \uparrow 2.8\%}}$ & \textbf{95.47}$_{\textcolor{ForestGreen}{\footnotesize \uparrow 5.5\%}}$ \\
\cmidrule(lr){2-16}
 & PixelSeal (128) & 0.805 & 37.54 & 0.719 & 20.34 & 0.683 & 19.89 & 0.789 & 34.95 & 0.754 & 30.59 & 0.561 & 2.60 & 0.750 & 28.96 \\
 & \quad + CAT & 0.806 & 37.84 & 0.728 & 21.93 & 0.726 & 23.98 & 0.796 & 36.62 & 0.778 & 33.62 & 0.651 & 10.65 & \textbf{0.768}$_{\textcolor{ForestGreen}{\footnotesize \uparrow 2.4\%}}$ & \textbf{31.37}$_{\textcolor{ForestGreen}{\footnotesize \uparrow 8.3\%}}$ \\
\midrule
\multirow{7}{*}{\shortstack{\textbf{SAV-Test}\\\textbf{(OOD)}}} & RivaGAN (32) & 0.941 & 25.13 & 0.740 & 13.02 & 0.776 & 13.66 & 0.812 & 14.94 & 0.868 & 19.17 & 0.502 & 2.05 & 0.811 & 15.68 \\
\cmidrule(lr){2-16}
\morecmidrules\cmidrule(lr){2-16}
 & VideoSeal 0.0 (96) & 0.945 & 68.05 & 0.914 & 57.84 & 0.834 & 43.75 & 0.933 & 65.09 & 0.897 & 57.69 & 0.716 & 16.33 & 0.903 & 58.21 \\
 & \quad + CAT & 0.952 & 70.46 & 0.908 & 57.43 & 0.889 & 53.14 & 0.923 & 63.18 & 0.926 & 63.30 & 0.794 & 29.14 & \textbf{0.913}$_{\textcolor{ForestGreen}{\footnotesize \uparrow 1.1\%}}$ & \textbf{59.97}$_{\textcolor{ForestGreen}{\footnotesize \uparrow 3.0\%}}$ \\
\cmidrule(lr){2-16}
 & VideoSeal 1.0 (256) & 0.851 & 101.68 & 0.823 & 87.24 & 0.739 & 63.77 & 0.842 & 100.10 & 0.804 & 87.90 & 0.599 & 9.54 & 0.811 & 88.22 \\
 & \quad + CAT & 0.841 & 95.59 & 0.827 & 89.13 & 0.808 & 82.13 & 0.834 & 94.25 & 0.832 & 93.66 & 0.752 & 55.46 & \textbf{0.826}$_{\textcolor{ForestGreen}{\footnotesize \uparrow 1.9\%}}$ & \textbf{90.53}$_{\textcolor{ForestGreen}{\footnotesize \uparrow 2.6\%}}$ \\
\cmidrule(lr){2-16}
 & PixelSeal (128) & 0.798 & 36.26 & 0.687 & 16.19 & 0.671 & 17.79 & 0.763 & 29.82 & 0.733 & 26.26 & 0.525 & 0.95 & 0.727 & 24.72 \\
 & \quad + CAT & 0.798 & 36.10 & 0.691 & 16.93 & 0.704 & 20.66 & 0.765 & 30.26 & 0.750 & 28.00 & 0.552 & 2.45 & \textbf{0.737}$_{\textcolor{ForestGreen}{\footnotesize \uparrow 1.5\%}}$ & \textbf{25.85}$_{\textcolor{ForestGreen}{\footnotesize \uparrow 4.6\%}}$ \\
\bottomrule
\end{tabular}
}
\label{tab:video_results_depth1}
\end{table}
\begin{table}[htbp]
\caption{\textbf{Under compositional video attacks, CAT delivers its strongest gains on the hardest combined and temporal attack pairs, with especially large improvements for PixelSeal.} Compositional video watermarking results on in-distribution Movie-Gen-Bench (ID) and out-of-distribution SA-V (OOD). Each cell reports bit accuracy and capacity for pairwise augmentation-family compositions, including combined corruption pairs such as Comp+Comp and temporally structured pairs such as Temp+Temp. For VideoSeal and PixelSeal, the `+ CAT` rows use Compositional Adversarial Training while the base rows use random augmentation training; higher is better for both metrics.}
\centering
\resizebox{0.96\textwidth}{!}{
\begin{tabular}{l l*{8}{cc}}
\toprule
 &  & \multicolumn{2}{c}{Val+Val} & \multicolumn{2}{c}{Val+Geom} & \multicolumn{2}{c}{Comp+Comp} & \multicolumn{2}{c}{Comp+Geom} & \multicolumn{2}{c}{Geom+Geom} & \multicolumn{2}{c}{Geom+Temp} & \multicolumn{2}{c}{Temp+Temp} & \multicolumn{2}{c}{Overall} \\
\cmidrule(lr){3-4}\cmidrule(lr){5-6}\cmidrule(lr){7-8}\cmidrule(lr){9-10}\cmidrule(lr){11-12}\cmidrule(lr){13-14}\cmidrule(lr){15-16}\cmidrule(lr){17-18}
 & Model (bits) & Bit acc. $(\uparrow)$ & Capacity $(\uparrow)$ & Bit acc. $(\uparrow)$ & Capacity $(\uparrow)$ & Bit acc. $(\uparrow)$ & Capacity $(\uparrow)$ & Bit acc. $(\uparrow)$ & Capacity $(\uparrow)$ & Bit acc. $(\uparrow)$ & Capacity $(\uparrow)$ & Bit acc. $(\uparrow)$ & Capacity $(\uparrow)$ & Bit acc. $(\uparrow)$ & Capacity $(\uparrow)$ & Bit acc. $(\uparrow)$ & Capacity $(\uparrow)$ \\
\midrule
\multirow{7}{*}{\shortstack{\textbf{Movie-Gen-}\\\textbf{Bench (ID)}}} & RivaGAN (32) & 0.701 & 8.57 & 0.740 & 10.03 & 0.715 & 9.03 & 0.740 & 10.01 & 0.772 & 11.05 & 0.820 & 14.54 & 0.873 & 18.98 & 0.758 & 11.39 \\
\cmidrule(lr){2-18}
\morecmidrules\cmidrule(lr){2-18}
 & VideoSeal 0.0 (96) & 0.947 & 69.20 & 0.978 & 82.26 & 0.825 & 40.55 & 0.975 & 81.26 & 0.961 & 76.52 & 0.978 & 82.34 & 0.904 & 60.87 & 0.967 & 79.19 \\
 & \quad + CAT & 0.937 & 66.09 & 0.973 & 80.01 & 0.937 & 65.17 & 0.974 & 80.32 & 0.960 & 75.96 & 0.976 & 81.39 & 0.960 & 74.65 & \textbf{0.972}$_{\textcolor{ForestGreen}{\footnotesize \uparrow 0.5\%}}$ & \textbf{79.52}$_{\textcolor{ForestGreen}{\footnotesize \uparrow 0.4\%}}$ \\
\cmidrule(lr){2-18}
 & VideoSeal 1.0 (256) & 0.823 & 87.28 & 0.869 & 113.11 & 0.658 & 33.70 & 0.867 & 112.26 & 0.839 & 99.39 & 0.869 & 113.42 & 0.749 & 66.80 & 0.853 & \textbf{106.90} \\
 & \quad + CAT & 0.829 & 89.90 & 0.860 & 107.49 & 0.809 & 78.77 & 0.857 & 105.42 & 0.835 & 95.19 & 0.859 & 106.48 & 0.835 & 93.55 & \textbf{0.855}$_{\textcolor{ForestGreen}{\footnotesize \uparrow 0.3\%}}$ & 104.51$_{\textcolor{red_down_arrow}{\footnotesize \downarrow 2.2\%}}$ \\
\cmidrule(lr){2-18}
 & PixelSeal (128) & 0.926 & 82.75 & 0.979 & 111.32 & 0.684 & 26.08 & 0.974 & 108.33 & 0.951 & 100.14 & 0.981 & 112.26 & 0.812 & 64.48 & 0.956 & 104.64 \\
 & \quad + CAT & 0.929 & 85.54 & 0.978 & 111.36 & 0.935 & 87.67 & 0.975 & 109.22 & 0.959 & 102.94 & 0.982 & 113.45 & 0.964 & 103.51 & \textbf{0.976}$_{\textcolor{ForestGreen}{\footnotesize \uparrow 2.1\%}}$ & \textbf{110.10}$_{\textcolor{ForestGreen}{\footnotesize \uparrow 5.2\%}}$ \\
\midrule
\multirow{7}{*}{\shortstack{\textbf{SAV-Test}\\\textbf{(OOD)}}} & RivaGAN (32) & 0.757 & 12.04 & 0.804 & 14.42 & 0.765 & 12.57 & 0.791 & 13.93 & 0.844 & 16.31 & 0.886 & 19.91 & 0.929 & 23.94 & 0.816 & 15.71 \\
\cmidrule(lr){2-18}
\morecmidrules\cmidrule(lr){2-18}
 & VideoSeal 0.0 (96) & 0.921 & 61.62 & 0.960 & 75.47 & 0.777 & 32.79 & 0.957 & 74.38 & 0.939 & 69.05 & 0.959 & 75.16 & 0.866 & 53.00 & \textbf{0.947} & \textbf{72.21} \\
 & \quad + CAT & 0.905 & 56.81 & 0.958 & 73.96 & 0.877 & 51.74 & 0.953 & 72.46 & 0.935 & 67.30 & 0.956 & 73.34 & 0.920 & 63.76 & \textbf{0.951}$_{\textcolor{ForestGreen}{\footnotesize \uparrow 0.5\%}}$ & 72.13$_{\textcolor{red_down_arrow}{\footnotesize \downarrow 0.1\%}}$ \\
\cmidrule(lr){2-18}
 & VideoSeal 1.0 (256) & 0.823 & 87.24 & 0.864 & 110.45 & 0.656 & 33.32 & 0.862 & 109.44 & 0.834 & 95.91 & 0.864 & 110.56 & 0.744 & 65.24 & \textbf{0.848} & \textbf{104.30} \\
 & \quad + CAT & 0.827 & 89.13 & 0.850 & 101.99 & 0.782 & 69.44 & 0.848 & 101.16 & 0.827 & 91.11 & 0.849 & 101.94 & 0.814 & 85.39 & 0.844$_{\textcolor{red_down_arrow}{\footnotesize \downarrow 0.5\%}}$ & 99.05$_{\textcolor{red_down_arrow}{\footnotesize \downarrow 5.0\%}}$ \\
\cmidrule(lr){2-18}
 & PixelSeal (128) & 0.832 & 53.45 & 0.915 & 83.98 & 0.653 & 20.14 & 0.907 & 80.49 & 0.875 & 70.21 & 0.914 & 83.08 & 0.773 & 50.66 & 0.891 & 78.13 \\
 & \quad + CAT & 0.897 & 73.02 & 0.956 & 98.80 & 0.855 & 64.09 & 0.951 & 96.65 & 0.928 & 88.32 & 0.957 & 99.07 & 0.908 & 83.58 & \textbf{0.948}$_{\textcolor{ForestGreen}{\footnotesize \uparrow 6.4\%}}$ & \textbf{96.18}$_{\textcolor{ForestGreen}{\footnotesize \uparrow 23.1\%}}$ \\
\bottomrule
\end{tabular}
}
\label{tab:video_results_depth2}
\end{table}

\paragraph{Video training and evaluation protocol.}
For the additional video results, we initialize from the pretrained image watermarking model and finetune on video data from Movie-Gen-Bench. We then evaluate the resulting video model on 100 evaluation videos from Movie-Gen-Bench for in-distribution testing and 100 evaluation videos from SA-V for out-of-distribution testing. This protocol isolates the effect of transferring the image-trained model into the video setting while keeping the downstream robustness evaluation fixed across the two benchmarks. As a baseline, we add RivaGAN, which is a Generative Adversarial Network based watermarking method \citep{zhang2019robust}.

\paragraph{CAT is most helpful on the hardest compositional and temporal video corruptions.}
Tables~\ref{tab:video_results_depth1} and~\ref{tab:video_results_depth2} extend the main results to video watermarking on in-distribution Movie-Gen-Bench \citep{polyak2024movie} and out-of-distribution SA-V \citep{ravi2024sam}. In the single-step setting, CAT improves overall capacity for all three learned backbones on both datasets: on Movie-Gen-Bench, overall capacity rises from 60.95 to 65.76 for VideoSeal 0.0, from 90.48 to 95.47 for VideoSeal 1.0, and from 28.96 to 31.37 for PixelSeal; on SA-V, it rises from 58.21 to 59.97, from 88.22 to 90.53, and from 24.72 to 25.85, respectively. In the compositional setting, the largest gains concentrate on the hardest combined and temporal compositions. For example, on Movie-Gen-Bench, VideoSeal 1.0 improves from 33.70 to 78.77 on Comp+Comp and from 66.80 to 93.55 on Temp+Temp, while PixelSeal improves from 26.08 to 87.67 on Comp+Comp and from 64.48 to 103.51 on Temp+Temp.

\paragraph{PixelSeal benefits consistently, while VideoSeal 1.0 is less stable in compositional OOD evaluation.}
PixelSeal shows the most reliable improvements across both ID and OOD video evaluation, with especially strong gains in the harder composed settings and corresponding increases in overall capacity. In the compositional OOD results on SA-V, PixelSeal improves from 78.13 to 96.18 overall capacity, with large gains on Comp+Comp (20.14 to 64.09), Comp+Geom (80.49 to 96.65), and Temp+Temp (50.66 to 83.58). VideoSeal 0.0 also benefits in the harder composed regimes, improving from 32.79 to 51.74 on Comp+Comp and from 53.00 to 63.76 on Temp+Temp on SA-V, even though its overall OOD capacity remains nearly unchanged (72.21 to 72.13). By contrast, VideoSeal 1.0 is less stable in the compositional OOD setting: although CAT improves hard regimes such as Comp+Comp (33.32 to 69.44) and Temp+Temp (65.24 to 85.39), its overall OOD performance decreases from 104.30 to 99.05 because the gains on those hard cases are offset by drops in easier families such as Val+Geom, Comp+Geom, and Geom+Temp.

Taken together, these results support the same conclusion as the image experiments: adaptive compositional adversaries are most useful when failure cases are rare, structured, and temporally or sequentially dependent. The strongest gains consistently appear on the hard combined and temporal video attacks that random augmentation is least likely to cover well, while the VideoSeal 1.0 compositional OOD results show that improvements on the hardest attack pairs do not automatically translate into better overall robustness if the learned attack policy becomes too narrow.
\section{Autoregressive Watermarking Results}\label{sec:autoregressive-results}

\begin{figure}[t]
    \centering
    \captionsetup[subfigure]{justification=centering,singlelinecheck=false}
    \begin{subfigure}[t]{\linewidth}
        \centering
        \includegraphics[width=\linewidth]{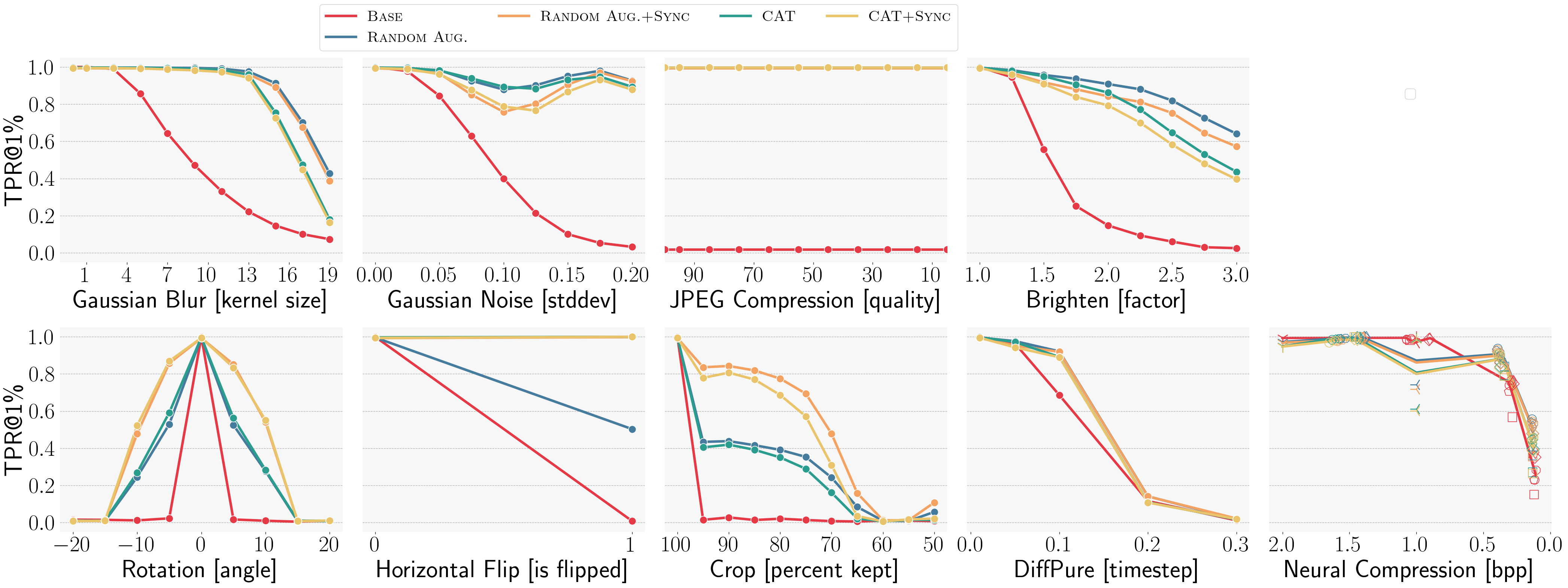}
        \caption{Taming}
    \end{subfigure}

    \medskip

    \begin{subfigure}[t]{\linewidth}
        \centering
        \includegraphics[width=\linewidth]{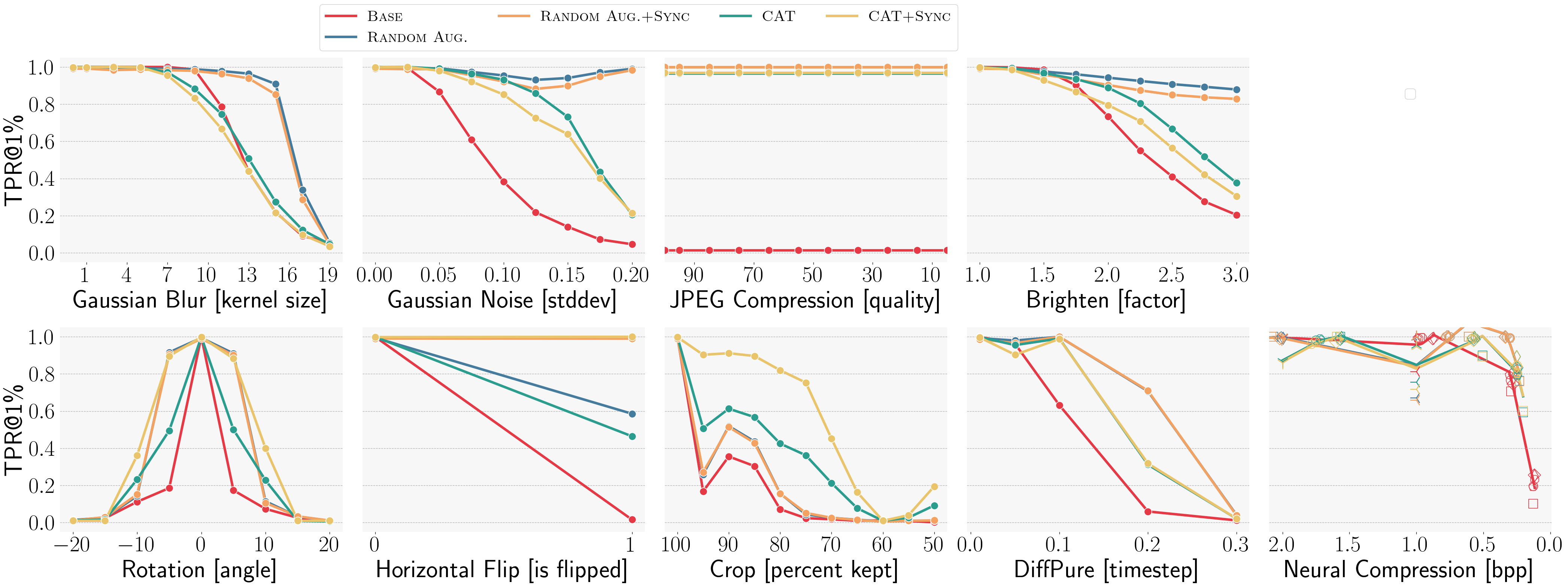}
        \caption{RAR-XL}
    \end{subfigure}
    \caption{\textbf{Continuous attack sweeps for autoregressive watermark robustness.} Each panel plots TPR@FPR$=1\%$ as the corruption strength varies. CAT consistently enlarges the robustness envelope over the Base watermark, while synchronization is particularly helpful for alignment-disrupting transformations such as cropping and horizontal flips.}
    \label{fig:autoregressive_attack_sweeps}
\end{figure}

\begin{figure}[t]
    \centering
    \captionsetup[subfigure]{justification=centering,singlelinecheck=false}
    \begin{subfigure}[t]{0.49\linewidth}
        \centering
        \includegraphics[width=\linewidth]{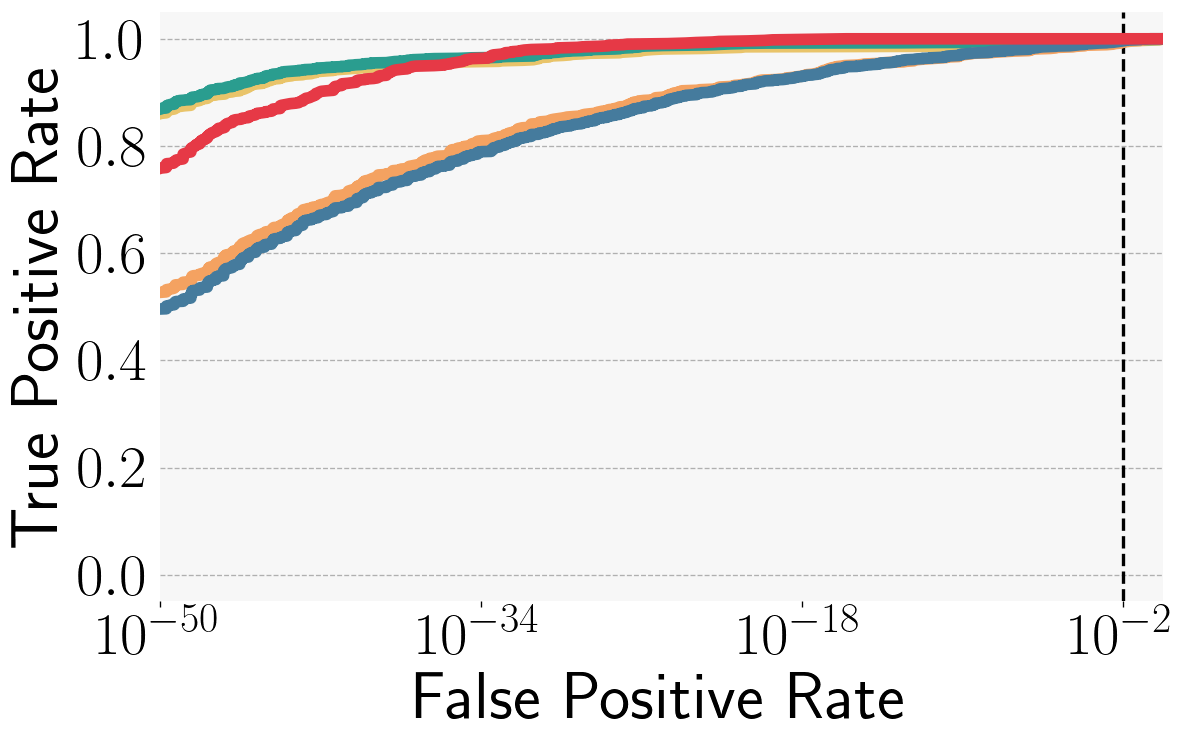}
        \caption{Taming}
    \end{subfigure}
    \hfill
    \begin{subfigure}[t]{0.49\linewidth}
        \centering
        \includegraphics[width=\linewidth]{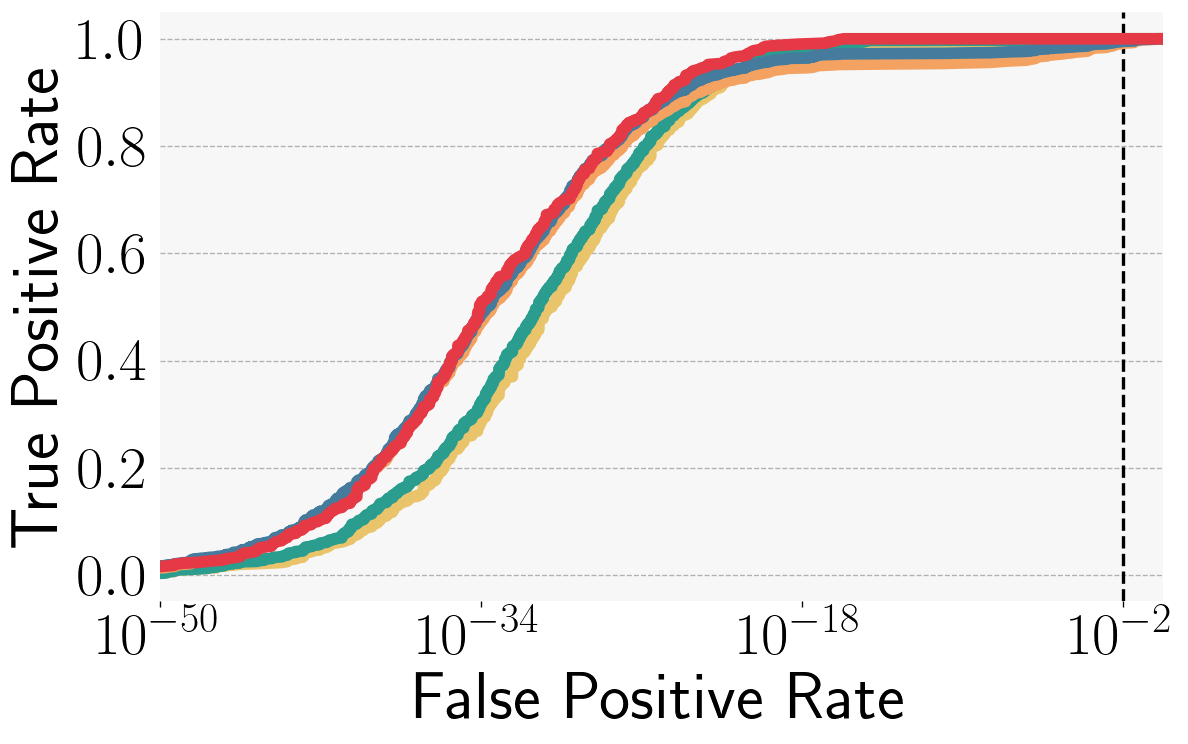}
        \caption{RAR-XL}
    \end{subfigure}
    \caption{\textbf{ROC curves for clean autoregressive watermark detection.} The dashed vertical line marks the operational false positive rate of $10^{-2}$ used in Tables~\ref{tab:autoregressive_taming} and~\ref{tab:autoregressive_rarxl}. Extreme low-FPR behavior depends on the backbone, but all methods approach near-perfect TPR at the operating point.}
    \label{fig:autoregressive_roc}
\end{figure}

\begin{figure}[t]
    \centering
    \captionsetup[subfigure]{justification=centering,singlelinecheck=false}
    \begin{subfigure}[t]{0.49\linewidth}
        \centering
        \includegraphics[width=\linewidth]{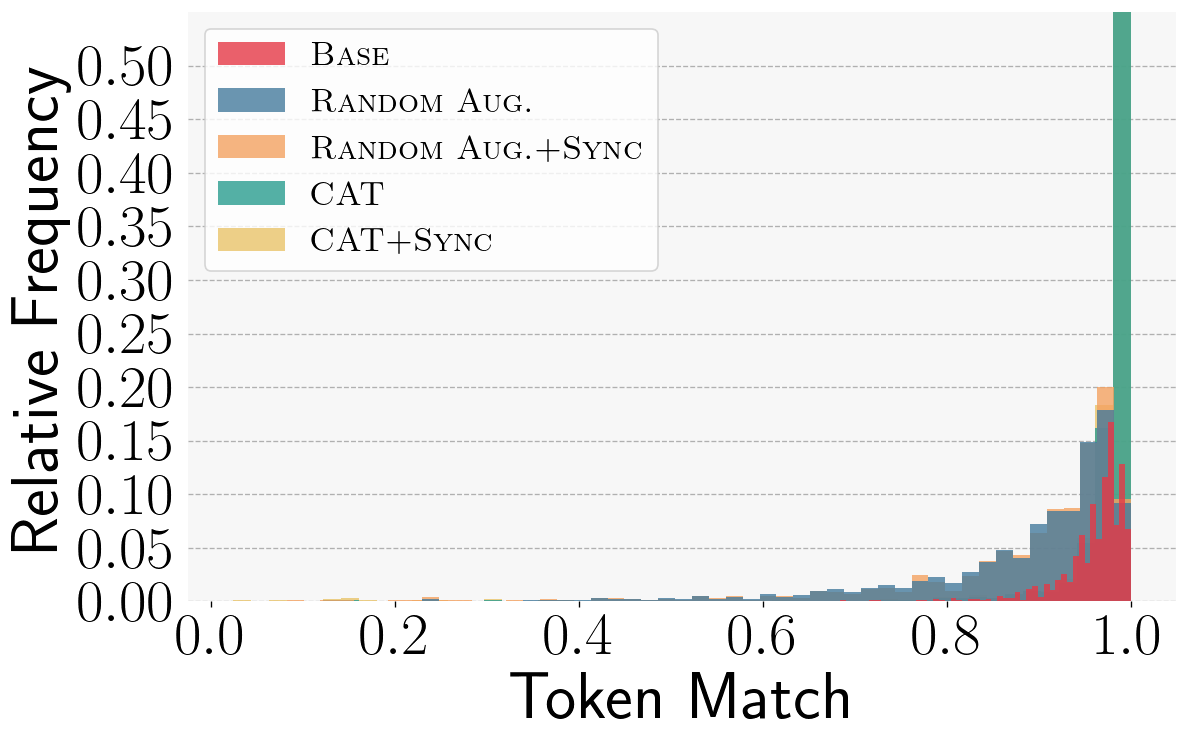}
        \caption{Taming}
    \end{subfigure}
    \hfill
    \begin{subfigure}[t]{0.49\linewidth}
        \centering
        \includegraphics[width=\linewidth]{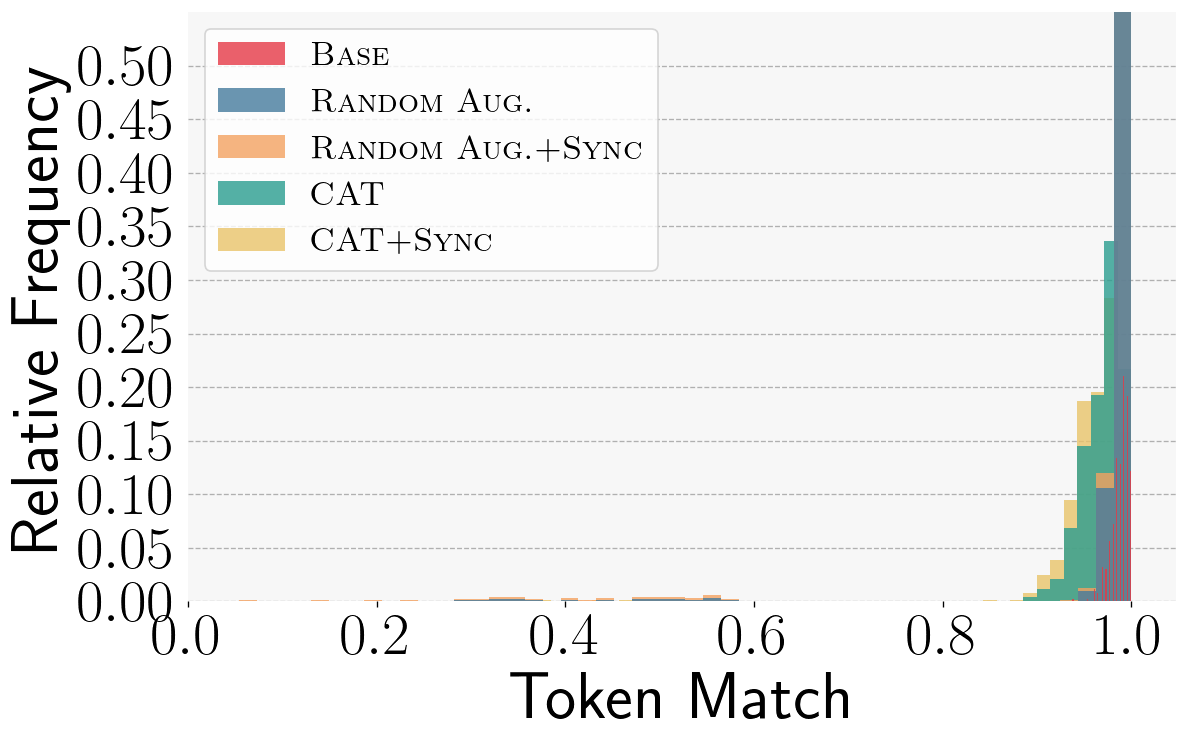}
        \caption{RAR-XL}
    \end{subfigure}
    \caption{\textbf{Token-match histograms under clean evaluation.} Synchronization broadens the high-match distribution and introduces a modest left tail, consistent with a detector that tolerates mild token misalignment after retokenization instead of requiring exact positional correspondence.}
    \label{fig:autoregressive_token_match}
\end{figure}

\paragraph{Training and evaluation protocol.}
We additionally evaluate CAT in the autoregressive image-generation setting using the WMAR framework of \citep{jovanovic2025watermarking}. Following their setup, the experiments are run on autoregressive generators based on Taming \citep{esser2021taming} and RAR-XL \citep{yu2025randomized}, and robustness is evaluated with the same detection metric reported in that work: true positive rate at a fixed false positive rate of $1\%$ (TPR@FPR$=1\%$). WMAR studies robustness under no attack, value perturbations, geometric perturbations, adversarial purification, and neural compression, and also introduces reverse cycle-consistency finetuning together with an optional synchronization layer to improve robustness after retokenization. Our tables therefore compare the no augmentation base watermark training, random-augmentation finetuning, random-augmentation finetuning with synchronization, CAT, and CAT with synchronization under the same attack families and detection metric. We supplement these aggregate benchmarks with three visual diagnostics: continuous attack sweeps in Figure~\ref{fig:autoregressive_attack_sweeps}, token-match histograms in Figure~\ref{fig:autoregressive_token_match}, and clean ROC curves in Figure~\ref{fig:autoregressive_roc}.

\paragraph{CAT improves the hardest geometric failure modes while maintaining performance on other attack families.}
Tables~\ref{tab:autoregressive_taming} and~\ref{tab:autoregressive_rarxl} show that the plain CAT variant consistently improves the hardest geometric failure modes while remaining competitive on the other attack families. On Taming, CAT raises geometric robustness from $0.01$ to $0.52$, while preserving strong performance on value perturbations ($0.94$), adversarial purification ($0.89$), and neural compression ($0.87$). On RAR-XL, the same pattern holds: CAT improves geometric robustness from $0.04$ to $0.35$ while still maintaining high detection on value perturbations ($0.92$), adversarial purification ($0.99$), and neural compression ($0.98$).

Figure~\ref{fig:autoregressive_attack_sweeps} makes the same pattern visible continuously across attack strengths. For both backbones, the Base watermark collapses quickly under rotation, cropping, and other geometric perturbations, whereas CAT substantially widens the usable operating range. On Taming, CAT maintains much higher detection than the Base watermark across blur, noise, brighten, and rotation even before synchronization is added. On RAR-XL, CAT is likewise far more robust than the Base watermark across the geometric stress tests, with especially clear gains throughout the crop sweep. Figure~\ref{fig:autoregressive_roc} shows that these robustness gains do not erase clean discriminability. The exact far-left-tail trade-off depends on the backbone---CAT and CAT+Sync are strongest at extremely small false positive rates on Taming, whereas the Base curve remains slightly sharper on RAR-XL---but all methods approach near-perfect true positive rates by the operational $10^{-2}$ threshold used in the tables.

\paragraph{Synchronization is what converts those gains into the strongest autoregressive robustness.}
The synchronization layer closes the remaining gap on geometric attacks: on both generators, CAT+Sync yields the strong geometric robustness, reaching $0.71$ on Taming and $0.72$ on RAR-XL, comfortably above both plain CAT and Random Augmentation Training. Figure~\ref{fig:autoregressive_attack_sweeps} shows where this extra gain comes from. On Taming, the synchronized variants are the only methods that remain reliable deep into the crop sweep, and they preserve perfect performance under horizontal flips. On RAR-XL, CAT+Sync is especially strong under aggressive cropping, clearly outperforming Random Aug.+Sync once the kept area becomes small, while both synchronized variants remain near-perfect under flips.

The token-match histograms in Figure~\ref{fig:autoregressive_token_match} visualize the mechanism behind this geometric resilience. Synchronization replaces a rigid concentration at almost-perfect token match with a broader high-match distribution and a mild left tail, most visibly on RAR-XL. This pattern suggests a decoder that tolerates small alignment mismatches instead of demanding exact token-wise correspondence after retokenization. At the same time, CAT without synchronization remains sharply concentrated near perfect token match on Taming but still lacks the crop and flip robustness of CAT+Sync, showing that stronger token identities alone are not enough; the model also needs a way to realign them after spatial shifts. Taken together, these results suggest that in the autoregressive setting CAT is most useful for hardening the token representations themselves, while synchronization is essential for converting that adversarial pressure into a watermark that stays decodable after geometric misalignment.

\clearpage

\begin{table}[htbp]
\centering
\caption{Autoregressive watermarking results for Taming-style generation, building on the WMAR framework of \citep{jovanovic2025watermarking}. Each entry reports TPR@FPR$=1\%$ under no attack, value perturbations, geometric perturbations, adversarial purification, and neural compression. Entries below $50\%$ are highlighted in red. Higher is better.}
\label{tab:autoregressive_taming}
\resizebox{0.82\textwidth}{!}{
\begin{tabular}{lccccc}
\toprule
Method & None & Value & Geometric & Adv. Purif. & Neural Comp. \\
\midrule
\textsc{Base} & 1.00 & \textcolor{red}{0.26} & \textcolor{red}{0.01} & 0.69 & 0.71 \\
\textsc{Random Aug.} & 1.00 & 0.94 & \textcolor{red}{0.38} & 0.92 & 0.90 \\
\textsc{Random Aug.+Sync} & 0.99 & 0.90 & 0.74 & 0.92 & 0.89 \\
\textsc{CAT} & 1.00 & 0.94 & 0.52 & 0.89 & 0.87 \\
\textsc{CAT+Sync} & 0.99 & 0.89 & 0.71 & 0.89 & 0.87 \\
\bottomrule
\end{tabular}
}
\end{table}

\begin{table}[htbp]
\centering
\caption{Autoregressive watermarking results for RAR-XL generation, building on the WMAR framework of \citep{jovanovic2025watermarking}. Each entry reports TPR@FPR$=1\%$ under no attack, value perturbations, geometric perturbations, adversarial purification, and neural compression. Entries below $50\%$ are highlighted in red. Higher is better.}
\label{tab:autoregressive_rarxl}
\resizebox{0.82\textwidth}{!}{
\begin{tabular}{lccccc}
\toprule
Method & None & Value & Geometric & Adv. Purif. & Neural Comp. \\
\midrule
\textsc{Base} & 1.00 & 0.53 & \textcolor{red}{0.04} & 0.63 & 0.77 \\
\textsc{Random Aug.} & 0.99 & 0.97 & \textcolor{red}{0.25} & 1.00 & 1.00 \\
\textsc{Random Aug.+Sync} & 0.99 & 0.95 & \textcolor{red}{0.38} & 1.00 & 1.00 \\
\textsc{CAT} & 1.00 & 0.92 & \textcolor{red}{0.35} & 0.99 & 0.98 \\
\textsc{CAT+Sync} & 1.00 & 0.86 & 0.72 & 0.99 & 0.97 \\
\bottomrule
\end{tabular}
}
\end{table}

\clearpage
\section{Forward and Backward Transfer}\label{sec:forward-backward-transfer}

\begin{figure}[t]
    \centering
    \includegraphics[width=\textwidth]{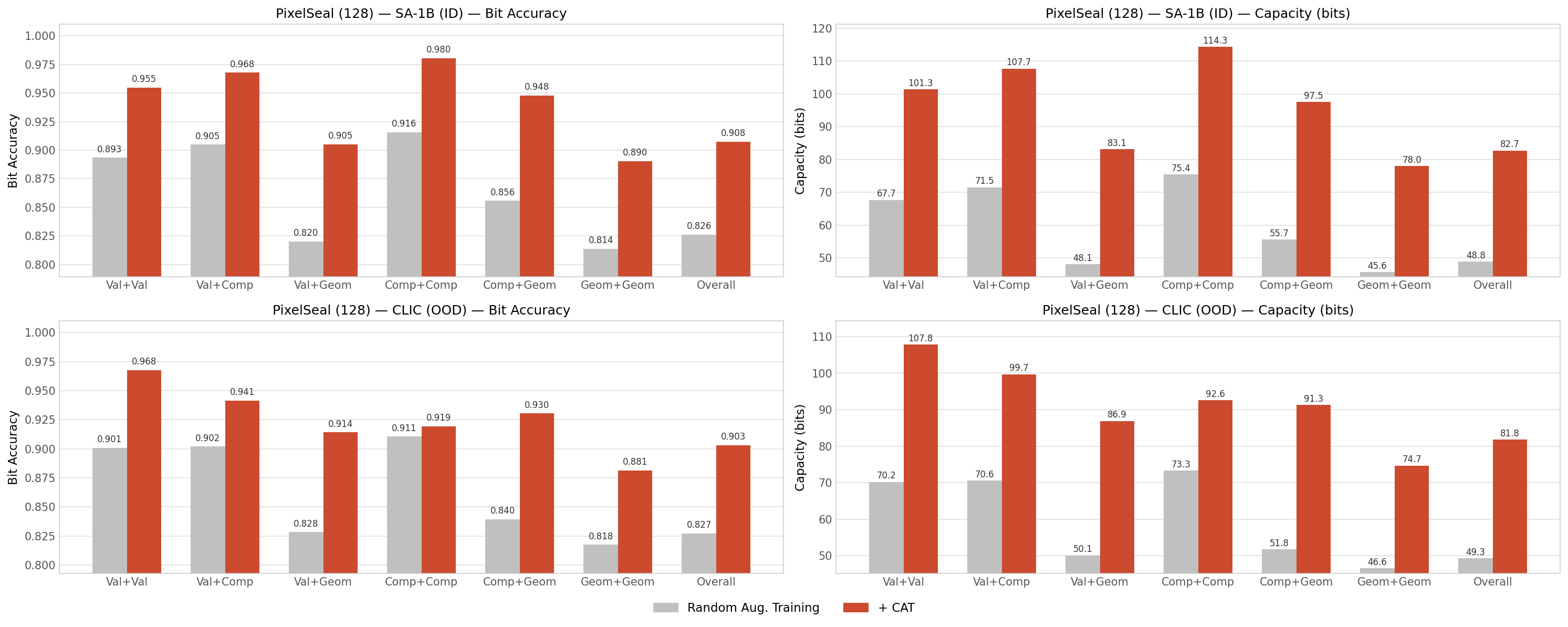}
    \caption{\textbf{Forward transfer from single-step training to two-step attack compositions.} Gray bars denote the random-augmentation training baseline and red bars denote CAT. CAT improves both bit accuracy and capacity across all reported composition families on SA-1B and CLIC, with the largest gains on the harder mixed and repeated compositions.}
    \label{fig:forward_transfer_summary}
\end{figure}

\begin{figure}[t]
    \centering
    \includegraphics[width=\textwidth]{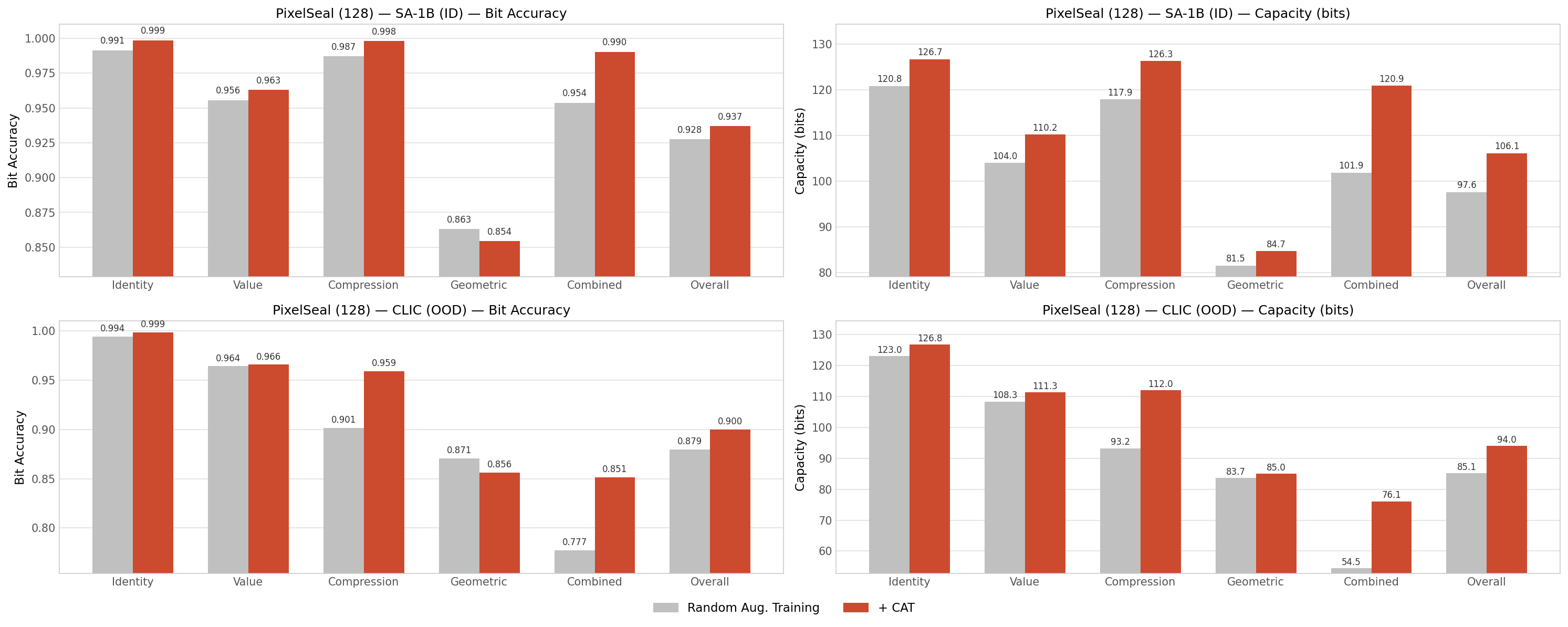}
    \caption{\textbf{Backward transfer from compositional training to single-step attacks.} Gray bars denote the random-augmentation training baseline and red bars denote CAT. Compositional CAT improves most single-step families and the overall metrics on both SA-1B and CLIC, while geometric bit accuracy dips slightly even as geometric capacity improves.}
    \label{fig:backward_transfer_summary}
\end{figure}

\paragraph{Transfer evaluation protocol.}
We additionally evaluate whether the robustness learned by CAT transfers across attack depths. In the \emph{forward transfer} setting, we take models trained with single-step attacks and test them on two-augmentation compositions. In the \emph{backward transfer} setting, we take models trained with compositional two-step attacks and test them on single augmentations. Figures~\ref{fig:forward_transfer_summary} and~\ref{fig:backward_transfer_summary} summarize PixelSeal results on in-distribution SA-1B and out-of-distribution CLIC using the same bit-accuracy and capacity metrics as the main image experiments.

\paragraph{Forward transfer shows that CAT learns robustness that generalizes to stronger, unseen attack compositions.}
Figure~\ref{fig:forward_transfer_summary} shows that the single-step CAT model substantially outperforms the single-step random-augmentation baseline when both are evaluated on two-step attack compositions. The gains are broad rather than isolated: on both SA-1B and CLIC, CAT improves both bit accuracy and capacity across every reported composition family, with especially large jumps in the harder mixed and repeated regimes. This pattern suggests that even when CAT is trained against only single-step adversaries, it still learns features that generalize to stronger composed perturbations.

\paragraph{Backward transfer remains strong, indicating limited over-specialization to composed attacks.}
Figure~\ref{fig:backward_transfer_summary} shows the converse evaluation: models trained on compositional attacks remain competitive when the test-time attack collapses back to a single augmentation. Across both SA-1B and CLIC, compositional CAT improves most single-step families and the overall metrics, while the only consistent regression is a small dip in geometric bit accuracy even though geometric capacity still increases slightly. Overall, the figure indicates that compositional CAT does not simply memorize multi-step attack structure; instead, it mostly learns robustness that transfers back to simpler single-step corruptions.

\clearpage
\section{Qualitative Results}
\label{sec:qualitative-results}

\begin{figure}[t]
    \centering
    \captionsetup{justification=centering,singlelinecheck=false}
    \includegraphics[width=\textwidth]{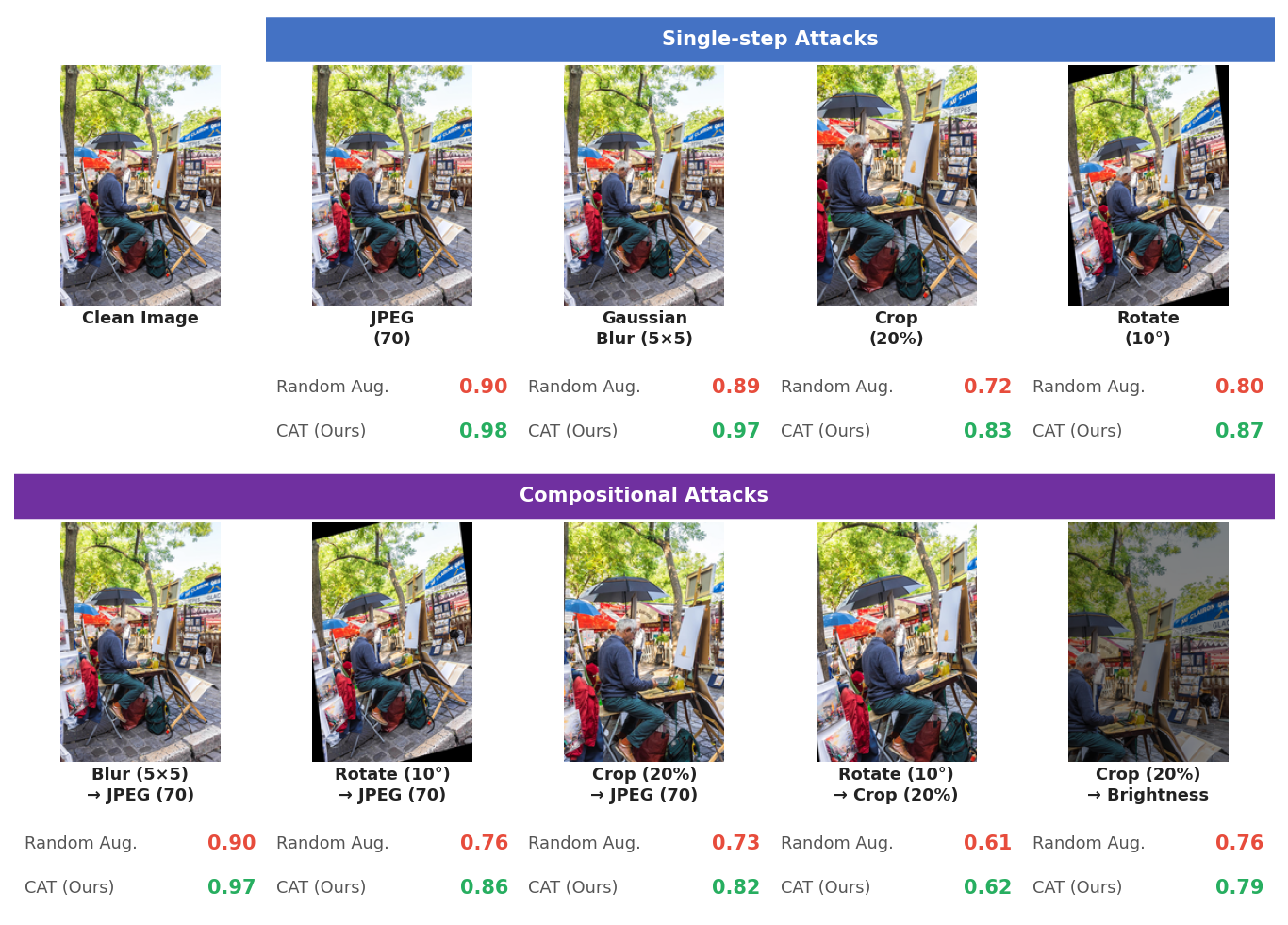}
    \caption{Additional qualitative results.}
    \label{fig:qualitative_results_1}
\end{figure}

\begin{figure}[t]
    \centering
    \captionsetup{justification=centering,singlelinecheck=false}
    \includegraphics[width=\textwidth]{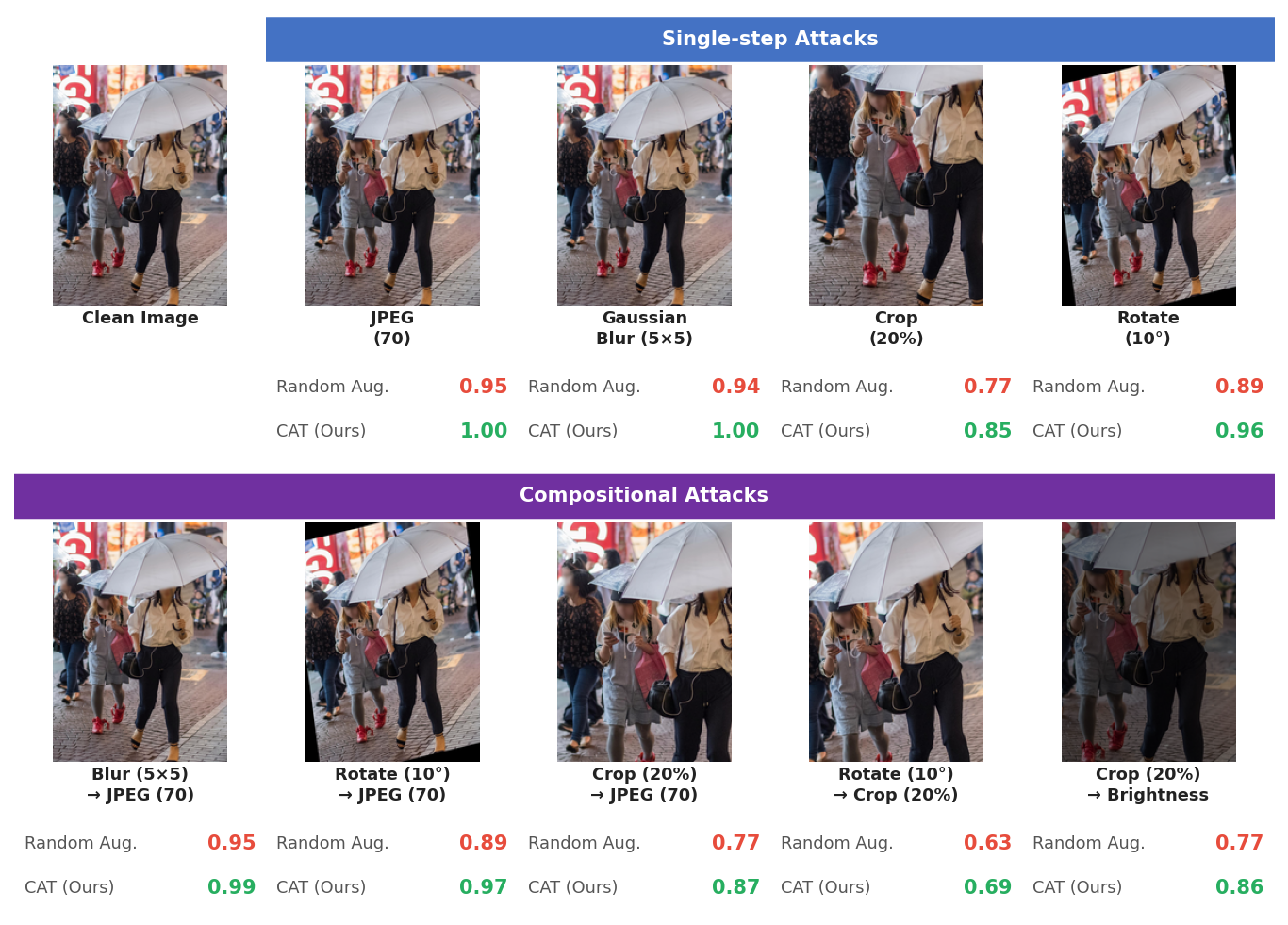}
    \caption{Additional qualitative results.}
    \label{fig:qualitative_results_2}
\end{figure}

\begin{figure}[t]
    \centering
    \captionsetup{justification=centering,singlelinecheck=false}
    \includegraphics[width=\textwidth]{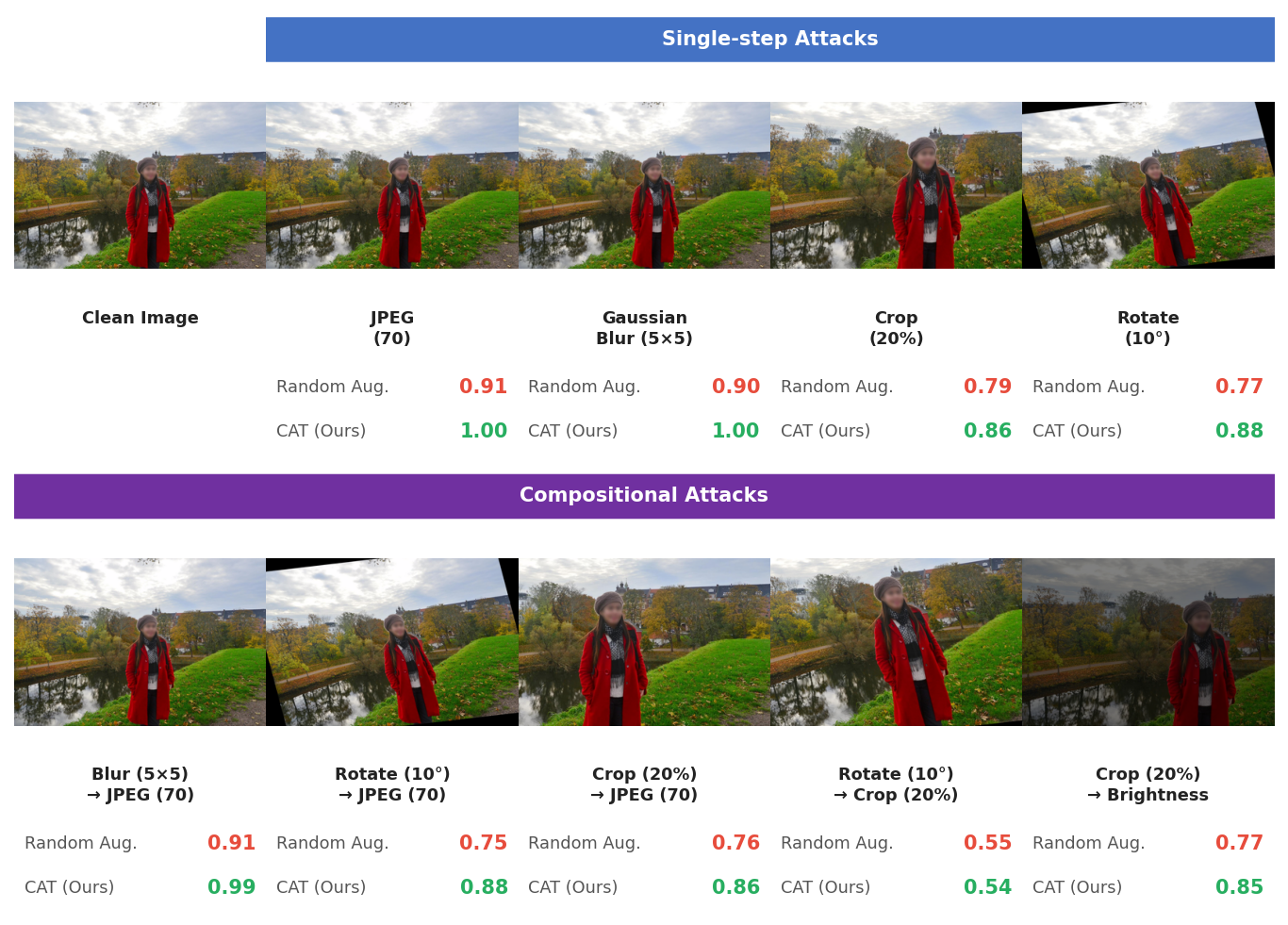}
    \caption{Additional qualitative results.}
    \label{fig:qualitative_results_3}
\end{figure}

\begin{figure}[t]
    \centering
    \captionsetup{justification=centering,singlelinecheck=false}
    \includegraphics[width=\textwidth]{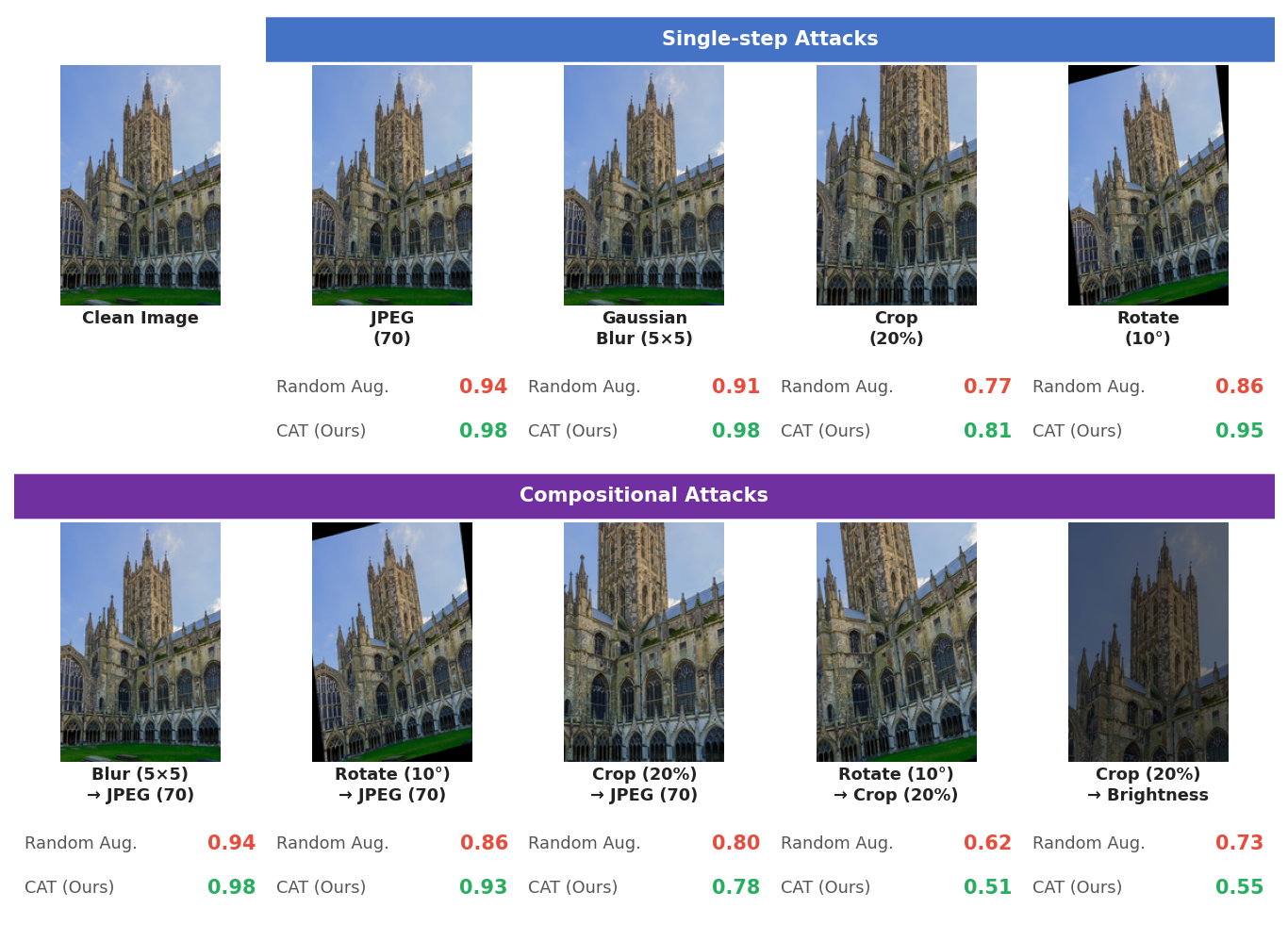}
    \caption{Additional qualitative results.}
    \label{fig:qualitative_results_4}
\end{figure}

\begin{figure}[t]
    \centering
    \captionsetup{justification=centering,singlelinecheck=false}
    \includegraphics[width=\textwidth]{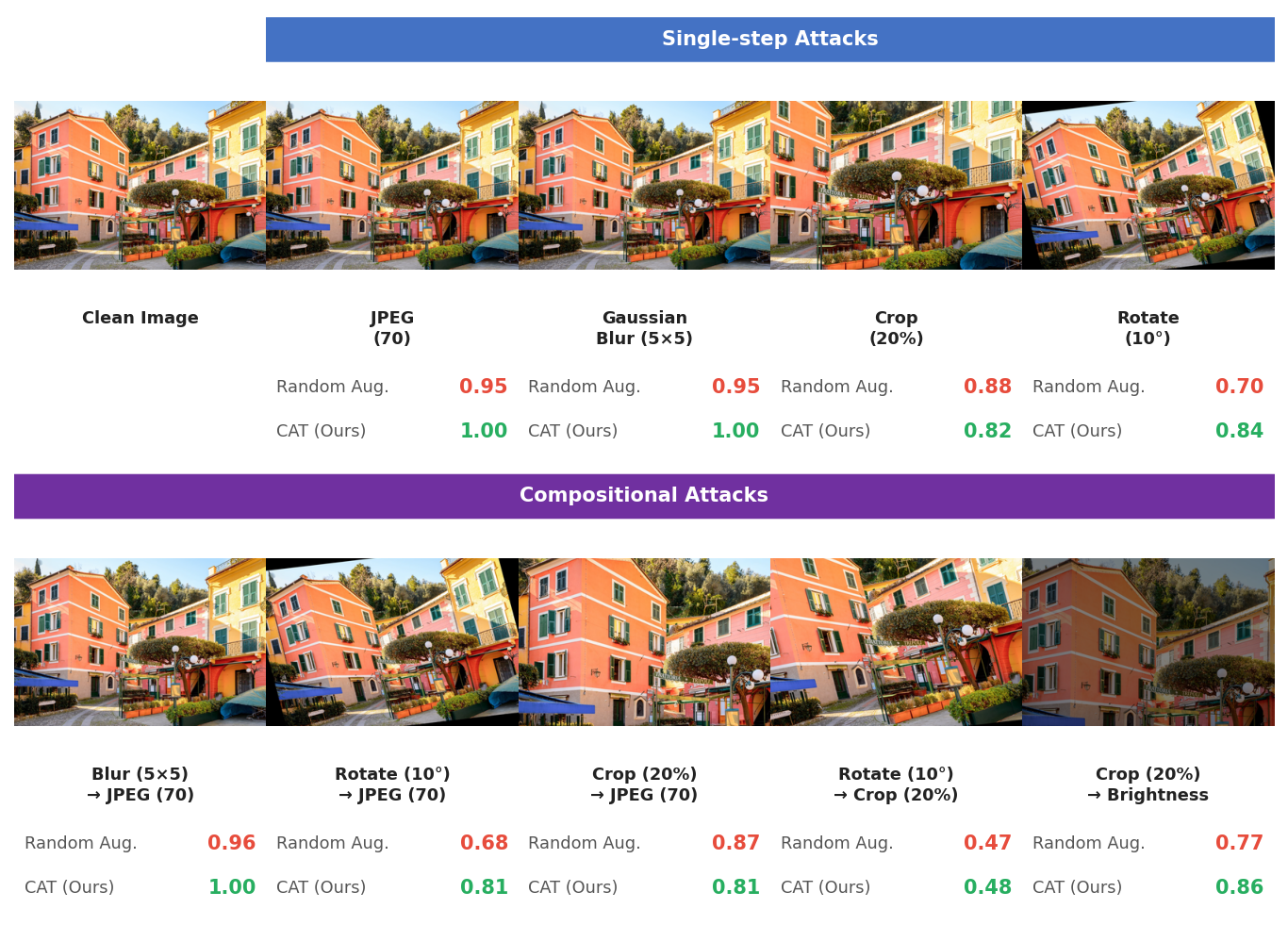}
    \caption{Additional qualitative results.}
    \label{fig:qualitative_results_5}
\end{figure}

Figures~\ref{fig:qualitative_results_1}--\ref{fig:qualitative_results_5} present additional qualitative examples for the image watermarking experiments discussed in the main paper.

\clearpage
\section{Broader Impact}
The work has clear positive applications in provenance tracking, creator attribution, and misuse detection for synthetic or redistributed media. At the same time, stronger watermarking methods can be used in restrictive content-control settings, and stronger attack models can also facilitate watermark-removal research. We believe that exposing brittle robustness claims before deployment is beneficial, but any release of code or trained adversaries should be accompanied by clear documentation of these dual-use risks and by usage restrictions that discourage misuse.
\end{document}